\documentclass[manuscript,screen]{acmart}
\usepackage{graphicx}
\usepackage{subfigure}
\usepackage[ruled]{algorithm2e}
\usepackage{multirow}
\usepackage{colortbl}
\usepackage{pifont}
\usepackage{makecell}

\makeatletter
\newcommand{\algorithmfootnote}[2][\footnotesize]{%
  \let\old@algocf@finish\@algocf@finish
  \def\@algocf@finish{\old@algocf@finish
    \leavevmode\rlap{\begin{minipage}{\linewidth}
    #1#2
    \end{minipage}}%
  }%
}
\makeatother

\definecolor{commentcolor}{RGB}{110,154,155}   

\newcommand{\PyComment}[1]{\ttfamily\textcolor{commentcolor}{\# #1}}  

\newcommand{\PyCode}[1]{\ttfamily\textcolor{black}{#1}} 

\AtBeginDocument{%
  }

\setcopyright{acmlicensed}
\copyrightyear{2018}
\acmYear{2018}
\acmDOI{XXXXXXX.XXXXXXX}
\acmConference[Conference acronym 'XX]{Make sure to enter the correct
  conference title from your rights confirmation email}{June 03--05,
  2018}{Woodstock, NY}
\acmISBN{978-1-4503-XXXX-X/2018/06}

\begin{document}

\title{WindowQuant: Mixed-Precision KV Cache Quantization based on Window-Level Similarity for VLMs Inference Optimization}

\author{Wei Tao}
\orcid{0009-0003-1262-5729}
\affiliation{%
  \institution{Huazhong University of Science and Technology}
  \city{Wuhan}
  \state{Hubei}
  \country{China}}

\author{Xiaoyang Qu}
\orcid{0000-0001-8353-4064}
\affiliation{%
  \institution{Ping An Technology (Shenzhen) Co., Ltd.}
  \city{Shenzhen}
  \state{Guangdong}
  \country{China}}

\author{Peiqiang Wang}
\orcid{0009-0009-4504-4721}
\affiliation{
\institution{Tsinghua University, Shenzhen International Graduate School}
\city{Shenzhen}
\state{Guangdong}
\country{China}
}

\author{Guokuan Li}
\orcid{0009-0005-7998-5520}
\affiliation{%
  \institution{Huazhong University of Science and Technology}
  \city{Wuhan}
  \state{Hubei}
  \country{China}}

\author{Jiguang Wan}
\orcid{0000-0003-3440-4460}
\affiliation{%
  \institution{Huazhong University of Science and Technology}
  \city{Wuhan}
  \state{Hubei}
  \country{China}}

\author{Kai Lu}
\orcid{0000-0002-7757-4083} 
\affiliation{%
  \institution{Huazhong University of Science and Technology}
  \city{Wuhan}
  \state{Hubei}
  \country{China}}
  
\author{Jianzong Wang}
\orcid{0000-0002-9237-4231}
\affiliation{%
  \institution{Ping An Technology (Shenzhen) Co., Ltd.}
  \city{Shenzhen}
  \state{Guangdong}
  \country{China}}

\renewcommand{\shortauthors}{Tao et al.}

\begin{abstract}
Recently, video language models (VLMs) have been applied in various fields. However, the visual token sequence of the VLM is too long, which may cause intolerant inference latency and GPU memory usage. Existing methods propose mixed-precision quantization to the key-value (KV) cache in VLMs based on token granularity, which is time-consuming in the search process and hardware inefficient during computation. This paper introduces a novel approach called WindowQuant, which employs window-adaptive mixed-precision quantization to optimize the KV cache. WindowQuant consists of two modules: window-level quantization search and window-level KV cache computation. Window-level quantization search quickly determines the optimal bit-width configuration of the KV cache windows based on the similarity scores between the corresponding visual token windows and the text prompt, maintaining the model accuracy. Furthermore, window-level KV cache computation reorders the KV cache windows before quantization, avoiding the hardware inefficiency caused by mixed-precision quantization in inference computation. Extensive experiments demonstrate that WindowQuant outperforms state-of-the-art VLM models and KV cache quantization methods on various datasets.
\end{abstract}



\received{20 February 2007}
\received[revised]{12 March 2009}
\received[accepted]{5 June 2009}

\maketitle

\section{Introduction}
\label{sec:introduction}

Large Language Models (LLMs) have gained significant popularity in both academia and
\begin{figure*}[]
    \centering
    \subfigure[]{
    \label{fig:example}
    \begin{minipage}[c]{0.3\linewidth}
        \centering
        \includegraphics[width=1\linewidth]{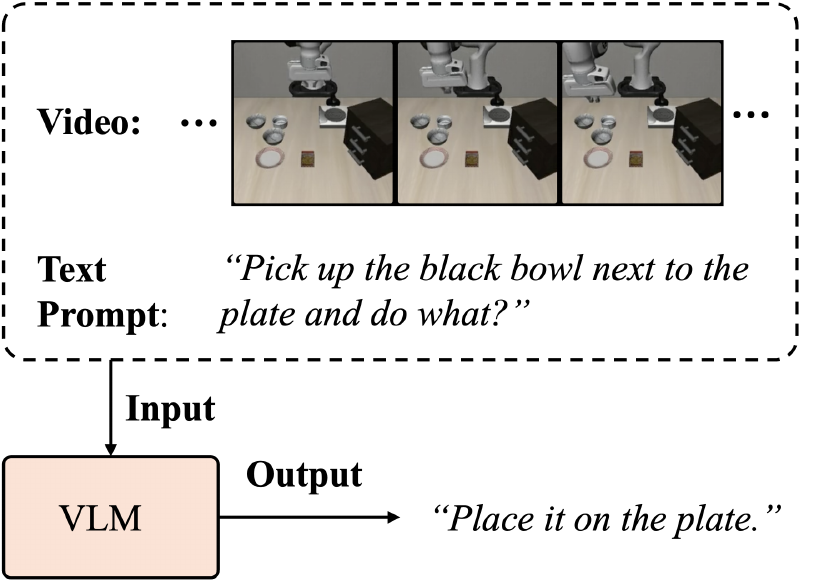}  
    \end{minipage}}
    \hspace{0.05\linewidth}
    \subfigure[]{
    \label{fig:challenge}
    \begin{minipage}[c]{0.6\linewidth}
        \centering
        \includegraphics[width=1\linewidth]{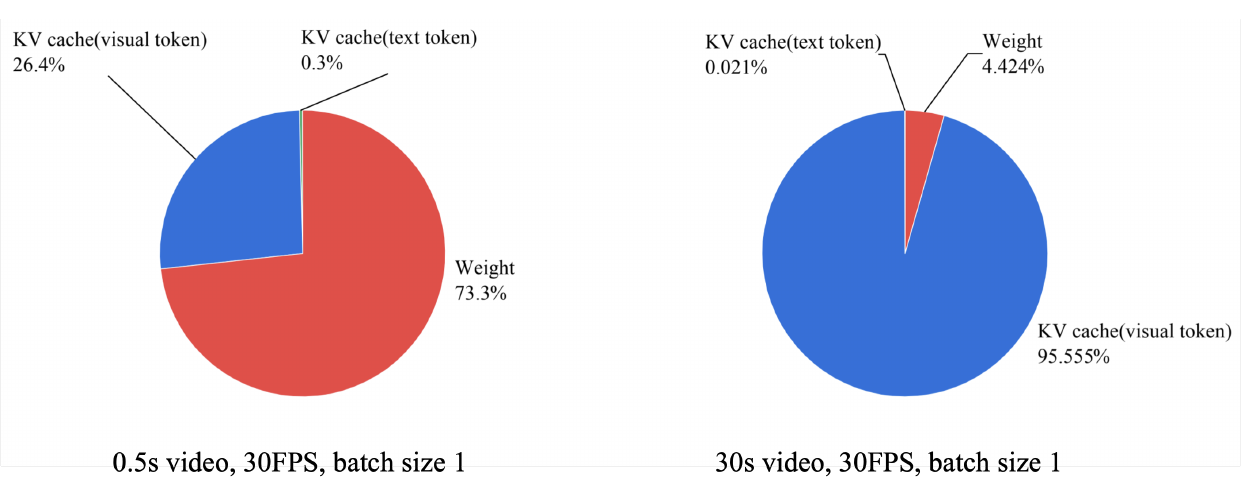}
    \end{minipage}}
    \caption{\textbf{(a)}: An example of VLM application. \textbf{(b)}: The memory usage comparison of KV cache (including visual token and text token) and weight of the Llava-OneVision-0.5B model when processing two videos of different lengths. Left: 0.5s video. Right: 30s video.}
    \Description{The example of VLM application and current challenge.}
\end{figure*}
industry due to their remarkable capabilities in understanding and generating human-like text. Recently, there has been a growing interest in extending LLMs to handle multimodal data, particularly video. Generally, when LLMs are used for video understanding, the input video is first converted into a sequence of frames, and each frame in the sequence is then transformed into tokens by spatiotemporal visual encoders, referred to as \textbf{visual tokens} (also called visual embeddings). Additionally, a piece of text is provided as a prompt to guide the LLM, and this text is also converted into tokens by text encoders, referred to as \textbf{text tokens} (also called text embeddings). The visual and text tokens are concatenated and fed into the LLM for processing. Therefore, LLMs for video understanding are also called Vision-Language Models (VLMs). VLMs have been applied to various fields such as video captioning, temporal grounding, action recognition, and embodied intelligence. Figure \ref{fig:example} demonstrates a typical example of a VLM application, where the VLM generates the answer to the question in the input text prompt based on the input video and the clue in the text prompt. By leveraging pretraining on large-scale video-text datasets, VLMs offer a powerful framework for reasoning about complex temporal dynamics and semantic content within videos.

\begin{figure}[t]
  \centering
  \begin{minipage}[t]{0.5\textwidth}
  \vspace{0pt}
    \centering
    \includegraphics[width=\linewidth]{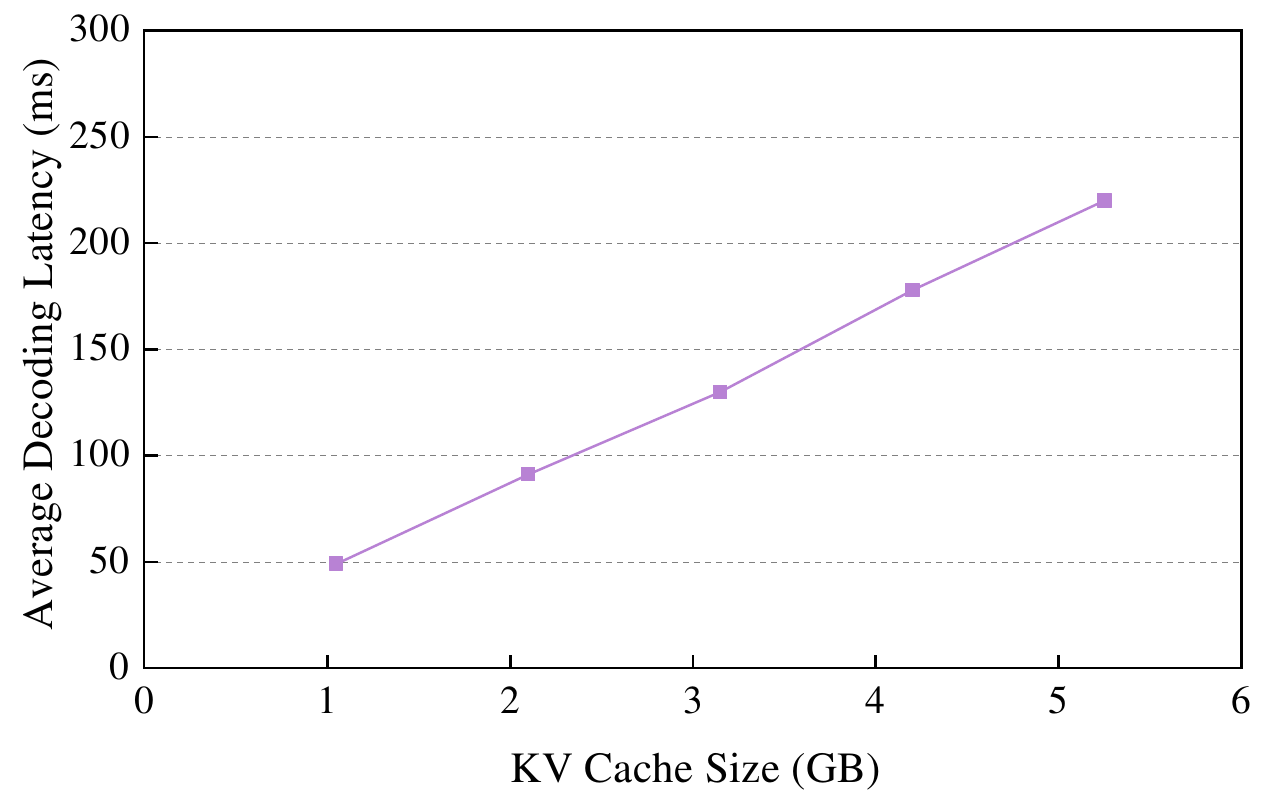}
    \caption{The average decoding latency of LLaVA-OneVision-Qwen2-7B model with different KV cache size.}
    \label{fig: kv cache latency}
    
  \end{minipage}
  \hspace{0.03\linewidth}
  \begin{minipage}[t]{0.45\textwidth}
  \vspace{0pt}
    \centering
    \includegraphics[width=\linewidth]{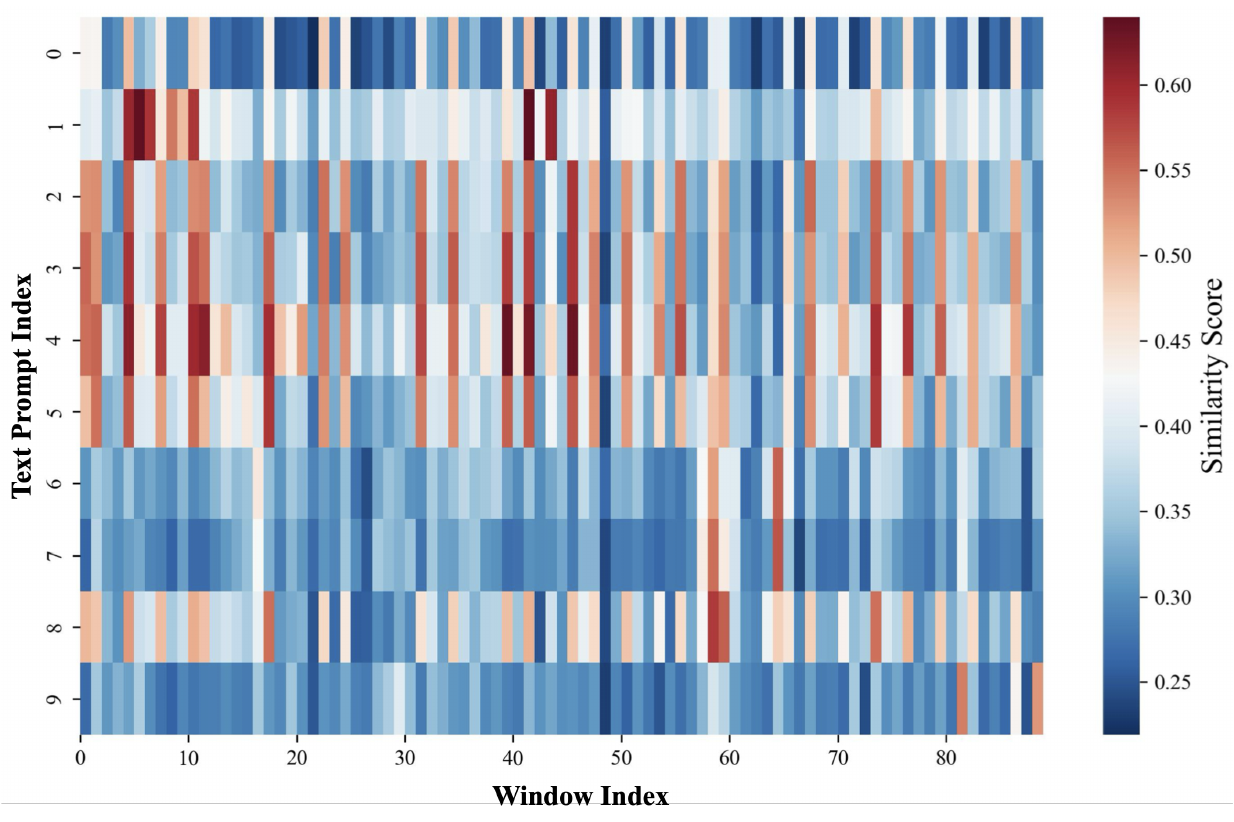}
    \caption{The similarity heatmap between a video and 10 text prompts. Most of the video windows are irrelevant to the text prompt.}
    \label{fig:vis}
  \end{minipage}
\end{figure}

However, current VLMs often face issues of insufficient GPU memory and excessively high inference latency when performing video understanding tasks. These challenges are mainly caused by the visual tokens' key-value (KV) cache. The memory footprint of a VLM's weights and text prompts remains static, but as the input video duration increases, the growing number of visual tokens leads to correspondingly larger KV cache memory consumption. For example, in the LLaVA-OneVision-0.5B model, the memory usage of the model weight is about 1GB. Generally, a text prompt contains dozens of words and will be transformed into several dozen text tokens, the memory usage of which is about 4MB. However, a single frame image will be converted into 256 visual tokens. Therefore, for a 30-second video at 30 FPS, even with a batch size of just 1 for inference, the memory consumption of the KV cache of visual tokens can reach an astonishing 21.6GB, easily exceeding the memory limits of typical GPUs. As illustrated in Figure \ref{fig:challenge}, when the input video is very short (only 0.5s), the model weight still dominates the memory usage. However, when the input video is a little longer (30s), the memory bottleneck shifts to the visual tokens' KV cache. The large KV cache not only leads to GPU memory insufficiency but also significantly increases inference latency. Since we need to transfer the KV cache data between GPU memory and the high-speed static random-access memory (SRAM) during inference, such a massive KV cache leads to substantial data transmission latency, ultimately increasing overall inference latency. Figure \ref{fig: kv cache latency} shows the variation of the average decoding latency required to generate one token for LLaVA-OneVision-Qwen2-7B under different KV cache sizes. As can be observed from the figure, the average decoding latency increases almost linearly with the growth of the KV cache size.


Researchers have proposed various methods to compress the KV cache, such as attention pruning \cite{shazeer2019fast,ainslie2023gqa,cai2024medusa}, removing or quantizing unimportant tokens \cite{zhang2024h2o, xiao2023efficient,han2023lm}, and considering optimizations from a system perspective \cite{dao2022flashattention, kwon2023efficient,dao2023flashattention}. Among them, the simplest and most effective approach is quantization. Quantization means converting the KV cache from the original FP16 type to a lower bit-width type (such as INT8, INT4, or even INT2). However, most existing KV cache quantization efforts \cite{liu2024kivi, sheng2023flexgen, zhao2024atom, lin2024qserve} uniformly quantize the KV cache to a single bit-width. In reality, the importance of tokens within the KV cache is hierarchical, meaning some tokens are more critical than others. Therefore, uniform quantization can easily cause accuracy degradation. Some papers \cite{kim2023squeezellm,yang2024no,hooper2024kvquant,dong2024qaq} have proposed implementing mixed-precision quantization based on the importance of the tokens to address this issue, but their quantization search strategies are token-level and very time-consuming. Even worse, mixed-precision quantization can cause hardware inefficiency during the inference computation, such as increased latency and memory usage.

In this paper, we propose to apply the mixed-precision quantization to the KV cache at a larger granularity, i.e., the \textbf{window level}.
We observed an interesting phenomenon: In the VLM inference process, not all visual tokens are relevant to the text prompt, and there is a significant amount of redundant information. We select a long video (which can be split into 89 windows) to simulate a long context and create 10 text prompts, calculating their similarity. As shown in Figure \ref{fig:vis}, for each text prompt, there is only a small portion of windows that are highly relevant, while most are irrelevant.
Since text prompts serve as the guiding instructions for VLMs, visual information unrelated to the text prompts can be regarded as contributing little to the task itself. Based on the above observation, we need to retain precision for the KV cache corresponding to the visual token windows that are highly relevant to the text prompt; otherwise, the model would lose critical information. For the KV cache corresponding to visual token windows irrelevant to the text prompt, quantizing them to very low bits will not significantly affect the model's accuracy. Given these observations, we propose a window-adaptive mixed-precision KV cache quantization method called \textbf{WindowQuant}.  

WindowQuant mainly contains two modules: window-level quantization search and window-level KV cache computation. Window-level quantization search calculates the similarity scores between the text prompt and each visual token window and determines the bit-width configuration of the KV cache windows corresponding to the visual token windows according to those scores. This module maintains the accuracy of VLMs quickly and effectively. Since our quantization configuration is determined from the window perspective—rather than assigning it token by token as in previous methods—our configuration search time is much shorter. Moreover, the subsequent quantization is also performed on a per-window basis, with all tokens in a window quantized together, which further reduces the overall quantization time. We further design the window-level KV cache computation module to avoid hardware inefficiency. Window-level KV cache computation reorders the context KV cache windows to ensure the same bit-width configuration in which KV cache windows are arranged contiguously in physical memory. This module significantly optimizes the inference latency and memory usage during VLM inference.

\begin{figure*}
    \centering
    \includegraphics[width=0.9\linewidth]{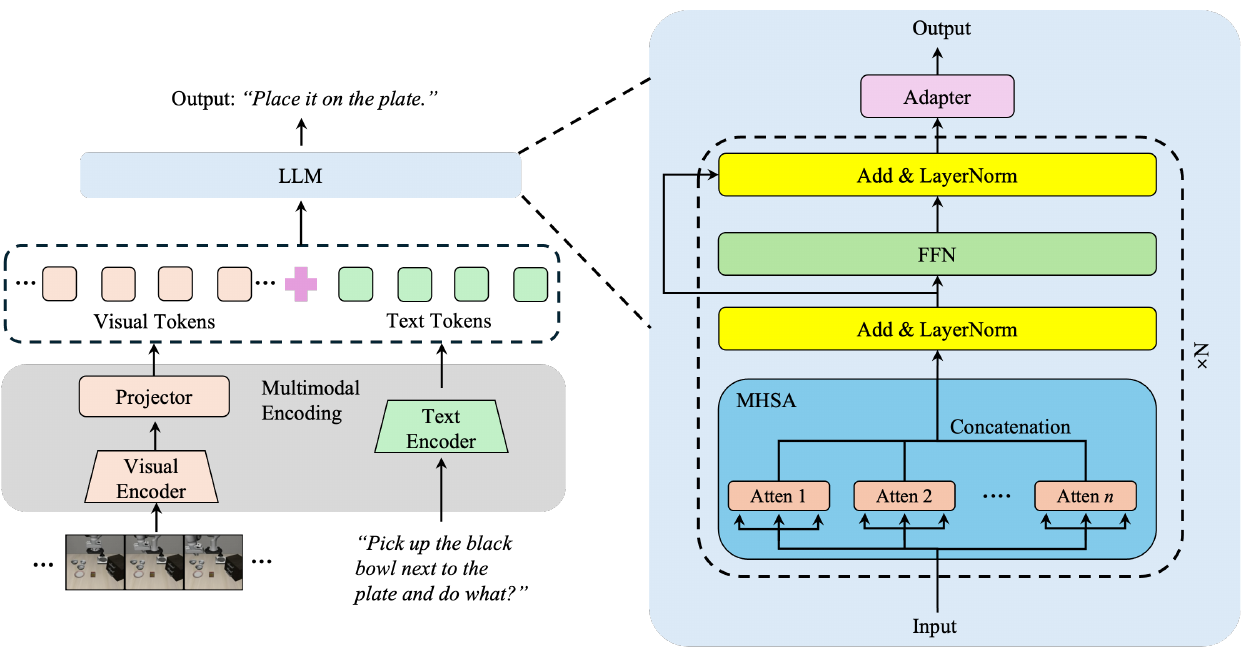}
    \caption{The architecture of typical VLMs.}
    
    \Description{Diagram showing the architecture of VLM.}
    \label{fig:architecture}
\end{figure*}

In summary, our contributions can be outlined as follows:
\begin{itemize}
\item We propose a window-adaptive mixed-precision KV cache quantization method called WindowQuant for effective and efficient VLM inference.
    \item WindowQuant contains the window-level quantization search module, which determines the bit-width configuration of KV cache windows based on the similarity scores between the corresponding visual token windows and the text prompt, maintaining the inference accuracy of the VLM.
    \item WindowQuant contains the window-level KV cache computation module, which reorders KV cache windows in different bit-width configurations before quantization, avoiding the issue of hardware inefficiency during inference caused by mixed-precision quantization.
    \item Extensive experiments validate that WindowQuant achieves better model accuracy, higher decoding throughput, and reduced memory usage compared with state-of-the-art (SOTA) methods.
\end{itemize}

\section{Preliminary}
In this section, we briefly review the necessary background and preliminaries to facilitate understanding of the proposed method, especially for readers who may be less familiar with vision–language model inference systems. The content presented here primarily summarizes established designs and observations from prior work, rather than introducing new technical contributions. The main content of this section is derived from the paper \textit{``Attention Is All You Need''} \cite{vaswani2017attention}.
\vspace{-2mm}

\subsection{VLM Architecture}

Figure \ref{fig:architecture} illustrates a typical VLM architecture \cite{NEURIPS2023_6dcf277e}. It primarily consists of a visual encoder, a text encoder, and an LLM. The input video and text prompt first go through a multimodal encoding stage. The video is transformed into visual tokens (also called visual embeddings) by the visual encoder and the projector (for modality alignment), while the text prompt is converted into text tokens (also called text embeddings) by the text encoder. These two kinds of tokens are then concatenated to form the input embeddings and fed into the LLM, which processes them to produce the output. Notably, both the visual encoder and text encoder are relatively lightweight, with the majority of computations occurring within the LLM, where the KV cache is also located. Therefore, our primary focus is on optimizing the LLM's computations, and to this end, we have also detailed the LLM's structural components. 

As illustrated in Figure \ref{fig:architecture}, the standard LLM  architecture is composed of multiple blocks stacked sequentially \cite{vaswani2017attention}. Each block typically consists of three core components: a Multi-Head Self-Attention (MHSA) module, a Feed-Forward Network (FFN) module, and a Layer Normalization module. 
These components operate in a cascaded manner, where each block takes the output features from the preceding block as input and applies transformations through its internal sub-modules.

The core of the LLM architecture is the self-attention mechanism employed in the MHSA block. Given a vector of input embeddings $\mathbf{X} = [x_1, x_2, ..., x_n]$, the MHSA module applies linear projections using a query weight matrix $\mathbf{W}^Q$, key weight matrix $\mathbf{W}^K$, and value weight matrix $\mathbf{W}^V$ to compute the query ($\mathbf{Q}$), key ($\mathbf{K}$), and value ($\mathbf{V}$) matrices. 
The calculation is illustrated in Equation \ref{eq:qkv}:
\begin{equation}
    \mathbf{Q}=\mathbf{XW}^{Q},\mathbf{K}=\mathbf{XW}^{K},\mathbf{V}=\mathbf{XW}^{V}
    \label{eq:qkv}
\end{equation}
The tuple ($\mathbf{Q}$,$\mathbf{K}$,$\mathbf{V}$) undergoes a self-attention operation, where the attention weights $\mathbf{A}$ are first computed by applying the text prompt matrix $\mathbf{Q}$ and the key matrix $\mathbf{K}$, as shown in Equation \ref{eq:attention weight}:
\begin{equation}
    \mathbf{A}=Attention(\mathbf{Q},\mathbf{K} )=\frac{\mathbf{Q} \mathbf{K}^T}{\sqrt{d_k}}
    \label{eq:attention weight}
\end{equation}
Here, $d_k$ denotes the dimensionality of the token embeddings. 
The attention weights are then processed and multiplied by the value matrix $\mathbf{V}_i$ to obtain the output feature $\mathbf{Z}$, as shown in Equation \ref{eq:softmax}:
\begin{equation}
    \mathbf{Z}_i=Softmax(\mathbf{A}_i+\mathbf{M})\mathbf{V}_i
    \label{eq:softmax}
\end{equation}
where the matrix $\mathbf{M}$ represents the mask matrix, which is used to ensure that the attention weights for tokens appearing after the current token in the sequence are set to zero. 
The mask matrix is only applied in the prefill stage. In the decoding stage, there is no mask matrix. 
We will introduce the prefill stage and decoding stage later.


\subsection{LLM Inference Process}

Figure \ref{fig:inference} illustrates the inference process of a typical pre-trained LLM in the VLM, where the blue boxes represent each output token. The inference process can be divided into two main stages:

\begin{figure*}
    \centering
    \includegraphics[width=0.8\linewidth]{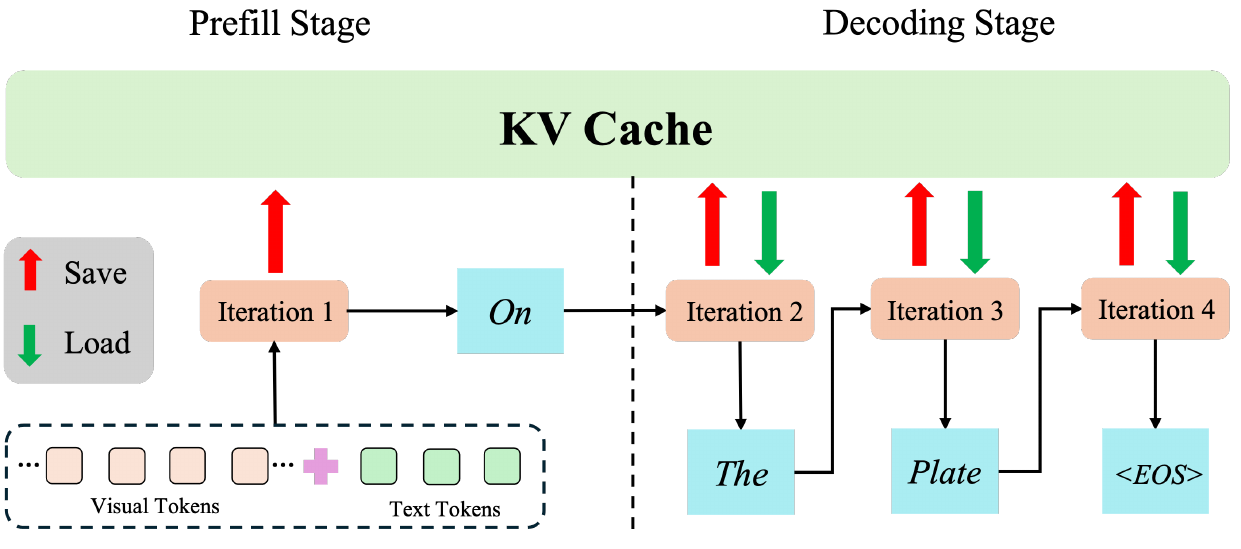}
    \caption{The inference process of typical LLMs.}
    
    \Description{Diagram showing the inference process of the VLM.}
    \vspace{-2mm}
    \label{fig:inference}
\end{figure*}

\begin{itemize}
    \item  \textbf{Prefill Stage}: During the prefill stage, the LLM receives the visual tokens and text tokens generated by the multimodal encoding process of the VLM. Then the VLM concatenates the visual tokens and text tokens and feeds them together into a pre-trained LLM. The pre-trained LLM calculates the input tokens in parallel and generates the first output token.
    
    \item \textbf{Decoding Stage}: During the decoding stage, the pre-trained LLM in the VLM combines the entire input token sequence with the already generated output tokens to produce the next token in the output sequence. This process iterates until the end-of-sentence token (<EOS>) is generated. 
\end{itemize}

The primary computational load in the aforementioned VLM inference process lies in the self-attention calculation mentioned in the previous section. Since the decoding stage is iterative and cannot be parallelized, it introduces the first major challenge in large model inference: each decoding step requires recomputing the K and V matrices for the entire input token sequence and the already generated output tokens, leading to significant redundant computation and severely increasing inference latency. To address this issue, researchers developed \textbf{KV cache}, which stores the K and V matrices of both the input tokens and the already generated output tokens in GPU memory. When computation is needed, these matrices are loaded from GPU memory into the GPU's high-speed SRAM. After computing the K and V matrices for a new output token, they are stored in the GPU memory and appended to the end of the existing KV cache. 

However, as discussed in Section \ref{sec:introduction}, an excessively large KV cache leads to a significant increase in GPU memory consumption and inference latency. Therefore, how to optimize the size of the KV cache has become a critical issue for VLM optimization. In the following, we introduce how to leverage mixed-precision quantization to reduce the KV cache size.

\section{Method}

In this section, we first outline the overall architecture of our WindowQuant approach. Next, we provide a detailed description of the two main components of WindowQuant, namely window-level quantization search and window-level KV cache computation.

\subsection{Architecture Overview}
\begin{figure*}[t]
    \centering
    \includegraphics[width=0.95\textwidth]{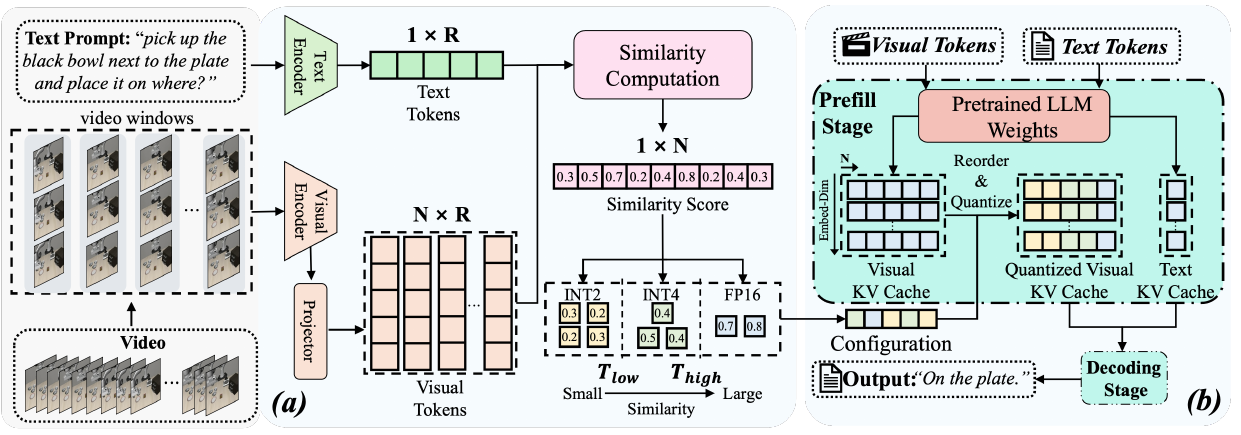}
    \caption{The architecture overview of WindowQuant. \textbf{(a)} The window-level quantization search module. \textbf{(b)} The window-level KV cache computation module.}
    \Description{Diagram showing the overview of our method.}
    \label{fig:system}
\end{figure*}

The architecture of WindowQuant is shown in Figure \ref{fig:system}. The input video is first segmented into several equal-length, short windows (If the length of the video is not divisible by the window size, we truncate the portion at the end of the video that cannot be divided by the window size. The KV cache of this portion will be kept in FP16 precision). The text prompt is converted into text tokens by the text encoder, while the video windows are converted into visual token windows by the visual encoder and the projector. Then, the window-level quantization search receives and processes the visual token windows and the text prompt, outputting the quantization bit-width configuration. Next, the visual token windows and the text tokens are fed into the pre-trained LLM for formal inference. During the prefill stage, we reorder the KV cache windows generated from the visual token windows. The reordered KV cache windows output from this module are subsequently quantized (all tokens in a window are quantized together) according to the configuration provided by the window-level quantization search module. Finally, after the decoding stage, the pre-trained LLM outputs the answer. To maintain model accuracy, we only quantize the KV cache of the input visual tokens while retaining FP16 precision for the KV cache of the output tokens and the text tokens. Given that, in general VLM inference tasks, the length of the visual tokens is significantly larger than that of the output tokens plus text tokens, we believe this strategy will not result in significant memory or inference latency overhead.

\subsection{Window-level Quantization Search}

\subsubsection{Quantization Configuration Determination}

Window-level quantization search draws inspiration from the concept of retrieval augmentation generation (RAG). RAG retrieves relevant documents or passages from an external knowledge source (such as a database or corpus) to augment the information contained in the current LLM input text prompt, thereby achieving better generation results. In our method, we find the video windows that are relevant to the text prompt and only retain the KV cache corresponding to them. To determine relevance, we directly use the visual token windows and the text tokens to calculate similarity. 

As shown in Figure \ref{fig:system}, video windows and text prompt are sent separately to the visual encoder and text encoder. After encoding and aligning modalities, we obtain the visual token windows and text tokens. Visual tokens in a window are essentially a subset of the original VLM visual tokens. Specifically, they consist of the tokens produced by the visual encoder from the video frames within the corresponding video window. We denote the original visual token matrix as
\begin{equation}
    \mathbf{V} = \{\mathbf{v}_1, \mathbf{v}_2, \ldots, \mathbf{v}_M\} \in \mathbb{R}^{M \times D}
\end{equation}

where $M$ is the number of visual tokens, $D$ is the dimension of encoder and $\mathbf{v}_i \in \mathbb{R}^D$ denotes the vector of the $i$-th visual token. For a video input consisting of $T$ frames, the total number of visual tokens is
\begin{equation}
    M = T \times K
\end{equation}
where $K$ denotes the number of visual tokens per frame, which is typically set to $256$ or $576$ depending on the input resolution and the configuration of the visual encoder.

\begin{figure*}
    \centering
    \includegraphics[width=1\linewidth]{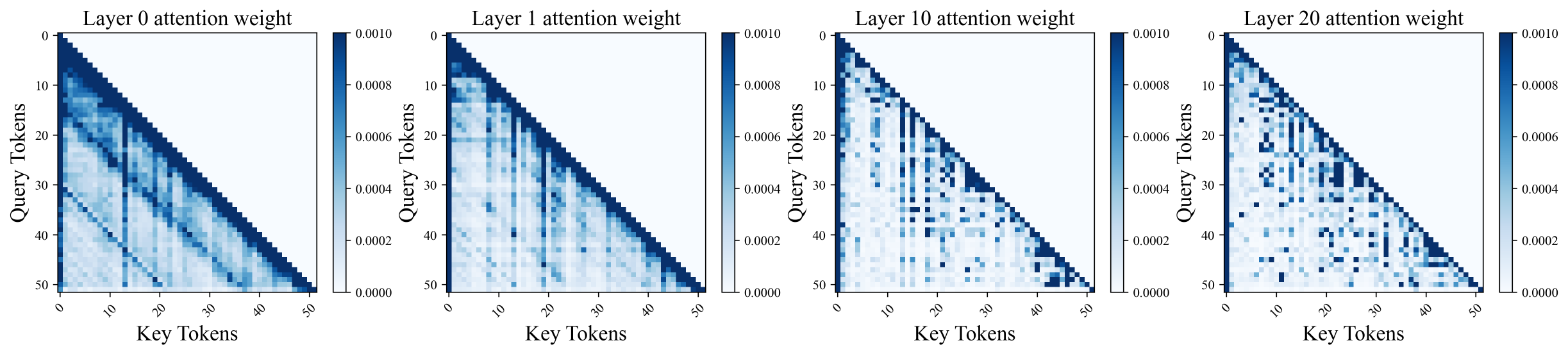}
    \caption{The attention weights of the tokens in four different layers in the pre-trained LLM of the Llava-Onevision-Qwen2-7B model.}
    \Description{Diagram showing an experimental result of the attention weights of the initial few tokens in LLMs.}
    \label{fig:first_window}
\end{figure*}

Similarly, we denote the text token matrix as
\begin{equation}
    \mathbf{T} = \{\mathbf{t}_1, \mathbf{t}_2, \ldots, \mathbf{t}_N\} \in \mathbb{R}^{N \times D}
\end{equation}
where $N$ is the number of text tokens and $\mathbf{t}_j \in \mathbb{R}^D$ denotes the vector of the $j$-th text token. Here, $D$ denotes the embedding dimension.

Suppose the window size is $S$, then the visual tokens matrix of the $i$-th window can be denoted as 
\begin{equation}
    \mathbf{W_i} = \{\mathbf{w}_{1}, \mathbf{w}_{2}, \ldots, \mathbf{w}_{S}\} = \{\mathbf{v}_{i\times S}, \mathbf{v}_{i\times S +1}, \ldots, \mathbf{v}_{i\times S+S-1}\} \in \mathbb{R}^{S\times D}
\end{equation}
where $\mathbf{w}_k \in \mathbb{R}^D$ denotes the vector of the $j$-th visual token in the window and there are $\lfloor{\frac{M}{S}\rfloor}$ windows in total. 

Next, we calculate the similarity between the text token matrix and each visual token window matrix, resulting in a similarity score list. The similarity formula is:
\begin{equation}
    sim(\mathbf{T}, \mathbf{W_i}) = \frac{1}{SN}\sum_{j=1}^{N}\sum_{k=1}^{S}\frac{\mathbf{t}_j \cdot \mathbf{w}_k}{||\mathbf{t}_j|| \times ||\mathbf{w}_k||} \quad i = 1,2,...,\lfloor{\frac{M}{S}\rfloor}
\end{equation}
where $||\cdot||$ means the L2 norm.

Subsequently, we set two thresholds, $T_{low}$ and $T_{high}$ ($0$ \textless\ $T_{low}$ \textless\ $T_{high}$ \textless $1$). We compare the similarity score at each index in the similarity score list with these two thresholds. For a similarity score greater than $T_{high}$, the visual token window at that index is considered highly similar to the text tokens, and the video window at that index is considered highly relevant to the text prompt, and we set the bit-width configuration of its corresponding KV cache window as FP16. For a similarity score less than $T_{low}$, the video window at that index is considered to have little relevance to the text prompt, and we set the bit-width configuration of its corresponding KV cache window as INT2. For a similarity score between $T_{low}$ and $T_{high}$, we adopt a compromise strategy, setting the bit-width configuration of its corresponding KV cache window as INT4. We will introduce how we set these two thresholds later.


\subsubsection{Fixed Precision of First Window}

Previous researchers \cite{xiao2023efficient} have demonstrated that the token at the initial position of an LLM plays a crucial role in the model’s performance, significantly influencing its accuracy. In our research, we further explore this phenomenon in VLMs. We conduct an experiment to depict the attention weights of the initial few visual tokens of different layers within the pre-trained LLM of Llava-OneVision-Qwen2-7B model(attention weights are the ``$A_i$" in Equation \ref{eq:attention weight}, which can reflect the contribution of a token to model inference accuracy). As depicted in Figure \ref{fig:first_window}, we observe that the attention weights for visual tokens at the initial positions are relatively higher than those for tokens in subsequent positions. This finding strongly suggests that visual tokens at the beginning of the sequence are highly influential, playing a critical role in determining the model's output. These initial visual tokens seem to capture essential contextual information, which is then propagated through the rest of the sequence.

In response to these observations, we propose to fix the precision of the first visual token window to FP16, regardless of its similarity score. This approach guarantees that the important information carried by the visual tokens at the start of the sequence won't be lost, thus protecting the model’s accuracy.

\subsubsection{Threshold Determination}

Note that LLMs contain numerous layers, and different layers may exhibit varying sensitivities to quantization. To investigate this, we follow the idea of SqueezeAttention \cite{wang2024squeezeattention}, using the cosine similarity between the hidden states before and after layer attention to represent the contribution of a layer to the model accuracy. Specifically, the similarity is calculated as follows:
\begin{equation}
    s_i = \frac{h_i \cdot h_i'}{||h_i||\cdot||h_i'||}
\end{equation}
where $h_i$ is the hidden states vector before the attention computation of layer $i$ and $h_i'$ is the hidden states vector after the attention computation of layer $i$.

A large $s_i$ indicates that the hidden states vector has not changed much when passing through the $i_{th}$ layer, which means that the contribution of this layer to the model accuracy is not high.

\begin{figure}[t]
    \centering
    \includegraphics[width=1\linewidth]{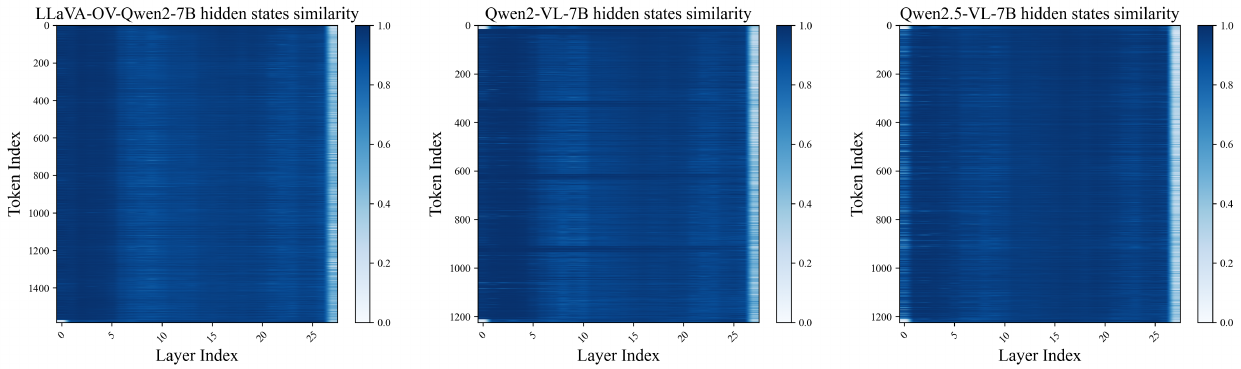}
    \caption{The hidden states similarity between layers, results are tested on three pre-trained LLMs of different VLM models. From left to right, the corresponding VLMs are LLaVA-QneVision-Qwen2-7B, Qwen2-VL-7B, and Qwen2.5-VL-7B (where ``OV'' in the figure denotes OneVision). }
    \label{fig:hidden sim}
\end{figure}

We depict the hidden states' similarity between layers on three pre-trained LLMs of different VLM models, and the results are shown in Figure \ref{fig:hidden sim}. In the figure, colors closer to white indicate lower similarity, while colors closer to blue indicate higher similarity. From the figure, we observe that across all three models, most layers exhibit high hidden states similarity, while only few final layers have low hidden states similarity. Therefore, most of the layers in the VLM have little contribution to the model accuracy. Only few layers are quite important to the model accuracy. Thus, we can adopt different quantization strategies for different layers in the VLM.

Specifically, we adopt the strategy of quantizing as many KV cache values as possible to INT2 for layers with large $s_i$. That is, we set both $T_{low}$ and $T_{high}$ relatively high, close to 1. Accordingly, we believe that layers with smaller $s_i$ contribute more to the model accuracy and are therefore not suitable for quantization. For these layers, we adopt the strategy of keeping as many KV cache values as possible in FP16 and setting both $T_{low}$ and $T_{high}$ relatively low, close to 0. For the other layers, we will quantize the values in the KV cache to INT4 as much as possible, meaning we will set a small $T_{low}$ close to 0 and a large $T_{high}$ close to 1. Finally, we design these two threshold functions to calculate the $T_{low}$ and $T_{high}$ of layer $i$:
\begin{equation}
    T_{low}^i = f_1(s_i)=\frac{e^{\alpha s_i}-1}{e^\alpha-1}
    \label{eq: f1}
\end{equation}
\begin{equation}
    T_{high}^i = f_2(s_i)=\frac{e^{-\alpha s_i}-1}{e^{-\alpha}-1}
    \label{eq: f2}
\end{equation}
where $\alpha$ is a pre-defined hyperparameter that controls the threshold. These two functions perfectly meet the aforementioned requirements. When \( s_i \) approaches 1, both \( f_1(s_i) \) and \( f_2(s_i) \) approach 1; when \( s_i \) approaches 0, both \( f_1(s_i) \) and \( f_2(s_i) \) approach 0. In other cases, \( f_1(s_i) \) tends to be closer to 0 while \( f_2(s_i) \) tends to be closer to 1. Moreover, within the interval [0, 1], \( f_1(s_i) \) is always less than or equal to \( f_2(s_i) \), perfectly satisfying the requirements for \( T_{\text{low}} \) and \( T_{\text{high}} \).

The overall algorithm of our window-level quantization search is shown in Algorithm \ref{alg1}.
\begin{algorithm}

    \caption{Pseudocode of window-level quantization search}
    \KwIn{Threshold parameter $a$, VLM model $M$, window size $S$, VLM visual encoder $VE$, VLM text encoder $TE$, video dataset $V$, text prompt $T$, Number of model layers $N$}
    \KwOut{Quantization bit-width configuration C}
    Sample a calibration dataset $V' = sample(V)$\;
    Run a prior inference with standard flash attention on the calibration dataset and achieve hidden states $H = M(V'+T)$ \;
    Initialize two threshold sets $T_{low}$, $T_{high}$ \; 
    \For{$i = 0$ to $N-1$}{
        $s_i = \frac{H_i \cdot H_{i+1}}{||H_i|| \cdot ||H_{i+1}||}$\;
        $T_{low}^i=\frac{e^{as_i}-1}{e^a-1}$\;
        $T_{high}^i=\frac{e^{-as_i}-1}{e^{-a}-1}$\;
    }
    Achieve visual embedding $E_v=VE(D)$ and text embedding $E_t=VT(T)$ \;
    Split visual embedding into windows $E_w = split(E_v, S)$\;
    \For{$i = 0$ to $N-1$}{
        \For{$j = 0$ to $len(E_w)-1$}{
            $sim_j^i = \frac{E_w^i \cdot E_t}{||E_w^i||\cdot||E_t||}$ \;
            \If{$sim_j^i > T_{high}^i$}{
                $C_j^i = 16$
            }
            \ElseIf{$sim_i^j < T_{low}^i$}{
                $C_j^i = 2$
            }
            \Else{
                $C_j^i = 4$
            }
        }
    }
    Return $C$\;
    \label{alg1}
\end{algorithm}

For batch-arriving requests, to preserve the convenience of batch computation, we apply the same quantization configuration to all requests within a batch. The procedure is as follows. First, for each request, $T_{low}$ and $T_{high}$ are identical, since they depend only on the current layer index and the hyperparameter $\alpha$. Second, for a given window position, we compute the similarity at that window for all requests in the batch, and determine the corresponding quantization configuration by comparing the similarity with $T_{low}$ and $T_{high}$. The quantization configuration that occurs most frequently is then selected and applied to that window for all requests in the batch. This process is repeated to determine the quantization configuration for all windows.

\subsection{Window-level KV Cache Computation}

\subsubsection{KV Cache Window Reordering}

Directly applying mixed-precision quantization can result in visual token windows with different bit-widths that are physically contiguous, which can lead to hardware inefficiency during inference computation. For example, modern GPUs use cache lines to optimize data reading, but data with different bit-widths cannot be aligned properly, potentially spanning multiple cache lines. This requires loading more cache lines, increasing the cache miss rate and memory access frequency, which significantly raises inference latency. Furthermore, if the hardware is designed to read multi-byte data simultaneously, such as single instruction multiple data (SIMD) architecture, some of the narrower bit-width data may only use part of the hardware resources, leading to GPU memory waste.

Therefore, we design the window-level KV cache computation module to address the hardware inefficiency problem. The algorithm pseudocode is shown in Algorithm \ref{alg:1} (For writing convenience, we omit the division by the scalar factor $\sqrt{d_k}$ in Equation \ref{eq:attention weight}). Specifically, during the prefill stage, we first calculate the output token. Then we adopt KV cache window reordering. Then, the KV cache windows with the same bit-width configuration are arranged together, making them contiguous in physical memory. Next, these windows are quantized following the bit-width configuration determined by the KV cache quantization search module. During the decoding stage, we multiply the $Q$ matrix by the transposed three blocks of the $K$ matrix with three different bit-widths to obtain three attention matrix blocks. These blocks are then concatenated along the last dimension to form the attention matrix. The attention matrix is then added to the mask matrix and processed with the softmax function. The processed attention matrix is divided into three blocks again, and each is multiplied by the $V$ matrix blocks with the corresponding bit-width to produce three small output matrices. The final output matrix is obtained by summing these three small output matrices.

\begin{algorithm}[ht]
\small
\caption{Pseudocode of window-level KV cache Computation in a Pytorch-like Style}
\algorithmfootnote{quant: the quantization function; fqm: FP16 matrix and quantized matrix multiply; mm: FP16 matrix multiply; cat: concatenation.}
\label{alg:1}

    \PyComment{s: the similarity score list}  \\
    \PyComment{N: the number of visual token windows} \\
    \PyComment{K, V: the K, V cache of the visual tokens} \\
    \PyComment{T\_low, T\_high: the two thresholds} \\
    
    \PyComment{\textbf{During the prefill stage:}} \\
    \PyCode{output = mm(softmax(mm(Q, K)), V)} \\
    \PyCode{for i in range(N):} \\
    \Indp   
        
        \PyCode{if s[i] < T\_low:} \\
        \Indp
            \PyCode{K\_int2 = torch.cat(K\_int2, K[i])}
            \\ 
            \PyCode{V\_int2 = torch.cat(V\_int2, V[i])}
            \\ 
        \Indm
        \PyCode{elif s[i] > T\_high:} \\
        \Indp
            \PyCode{K\_fp16 = torch.cat(K\_fp16, K[i])}
            \\ 
            \PyCode{V\_fp16 = torch.cat(V\_fp16, V[i])}
            \\ 
        \Indm
        \PyCode{else:} \\
        \Indp
            \PyCode{K\_int4 = torch.cat(K\_int4, K[i])}
            \\ 
            \PyCode{V\_int4 = torch.cat(V\_int4, V[i])}
            \\ 
        \Indm
    \Indm 
    K\_q2, K\_q4 = quant(K\_int2), quant(K\_int4) \\
    V\_q2, V\_q4 =  quant(V\_int2), quant(V\_int4) \\
    \PyComment{Q: the Q vector of the current token} \\
    \PyComment{len\_2, len\_4 = len(K\_q2), len(K\_q4)} \\
    \PyComment{\textbf{During the decoding stage:}} \\
    \PyCode{att = fqm(Q,transpose(K\_q2),2)} \\
    \PyCode{att = torch.cat(att,fqm(Q,transpose(K\_q4),4),-1)} \\
    \PyCode{att = torch.cat(att,mm(Q,transpose(K\_fp16)),-1)} \\
    \PyCode{att = softmax(att)} \\
    \PyCode{output = fqm(att[:len\_2)],V\_q2)} \\
     \PyCode{output += fqm(att[len\_2:len\_2+len\_4],V\_q4)} \\
     \PyCode{output += mm(att[len\_2+len\_4:],V\_fp16)} \\
     
\end{algorithm}

\subsubsection{Correctness Proof}
We prove that the output obtained in this way is the same as the output from the traditional computation method. In traditional computation method, assuming there are $N$ visual token windows, then $K$ matrix can be divided into a block matrix of the form $[K_1|K_2|...|K_N]^T$, and $V$ matrix can be divided into a block matrix of the form $[V_1|V_2|...|V_N]^T$, both of which are partitioned along the token length dimension. Obviously, attention matrix $A$ can also be divided into $[A_1|A_2|...|A_N]$. According to block matrix multiplication, the final output matrix $O$ is:
\begin{equation}
    O = [A_1V_1 + A_2V_2 + ... + A_NV_N]
\end{equation}

Now, if we perform KV cache window reordering, it is equivalent to changing the order of the sub-blocks within the $K$ and $V$ matrices. Suppose after reordering, $K$ becomes $[K_{x_1}|K_{x_2}|...|K_{x_N}]^T$, and $V$ becomes $[V_{x_1}|V_{x_2}|...|V_{x_N}]^T$, where $\{x_1, x_2, ..., x_N\}$ is a permutation of $\{1, 2, ..., N\}$. Since the softmax function is not affected by the order of the matrix sub-blocks, the softmax result equals the result obtained by first applying softmax to the original attention weights matrix and then performing reordering. That is, sub-blocks of matrix $A$ will eventually be arranged in the same order as the sub-blocks within $K$, which means $A$ is $[A_{x_1}|A_{x_2}|...|A_{x_N}]$. The final output $O'$ will be:
\begin{equation}
   O' =  [A_{x_1}V_{x_1} + A_{x_2}V_{x_2} + ... + A_{x_N}V_{x_N}]
\end{equation}
which is equal to the matrix $O$ due to the commutative invariance of matrix addition. 

\begin{figure*}[t]
    \centering
    \includegraphics[width=0.8\linewidth]{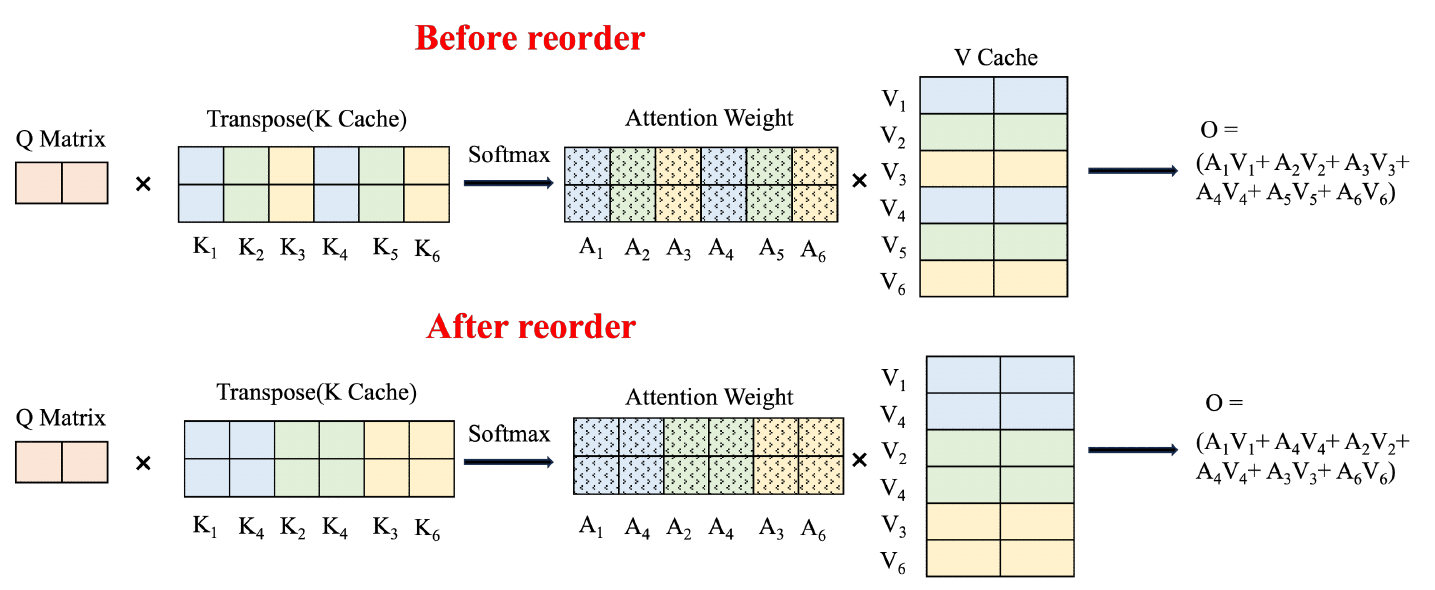}
    \caption{An example of KV cache window reordering.}
    \Description{Diagram showing an example of reordering.}
    \label{fig:reorder_example}
\end{figure*}

Figure \ref{fig:reorder_example} illustrates an example of the impact of KV cache window reordering on attention computation, where the number of windows is six. From the figure, we can see that, the result before reordering 
is $A_1V_1 + A_2V_2 + A_3V_3 + A_4V_4 + A_5V_5 + A_6V_6$, while the result after reordering is $A_1V_1 + A_3V_3 + A_2V_2 + A_4V_4 + A_3V_3 + A_6V_6$. Obviously, the attention computation result remains consistent before and after the KV cache window reordering.

\subsubsection{Implementation of Quantization}

We employ asymmetric round-to-nearest (RTN) quantization in our method. RTN quantization is a method that converts high-precision floating-point numbers into the closest representable integer in the target bit-width. Asymmetric quantization allows the quantized range to be offset (unlike symmetric quantization, where the range is centered around zero), enabling more precise representation of data with non-zero-centered or skewed distributions, which is well-suited for KV cache quantization. The formulas of the quantization and dequantization processes are as follows:

\textbf{Quantization:}
\begin{equation}
    Q = clamp(\lfloor \frac{x}{s} \rceil, q_{min}, q_{max})
\end{equation}

\textbf{Dequantization:}
\begin{equation}
    X = s \cdot (x - z)
\end{equation}
where $Q$ is the quantized tensor, $x$ is the original precision tensor, $X$ is the dequantized tensor, $s$ is the scaling factor, $z$ is the zero point, $q_{min}$ and $q_{max}$ are the integer range bounds ([0, $2^b$ - 1] for $b$-bit unsigned quantization and [-$2^{b-1}$, $2^{b-1}$-1] for $b$-bit signed quantization), $\lfloor \rceil$ is the round-to-nearest operator. $s$ and $z$ are calculated as follows:
\begin{equation}
    s = \frac{x_{max}-x_{min}}{q_{max}-q_{min}}, \quad
    z = q_{min} - \lfloor \frac{x}{s} \rceil
\end{equation}
where $x_{max}$ and $x_{min}$ are the max/min values in the original data.

We apply group-wise quantization to the KV cache, where each group's size corresponds to one window. The idea of group-wise quantization is derived from GPTQ \cite{frantar2022gptq}, which means all elements within the same group share uniform quantization parameters (i.e., scaling factor and zero point). Different regions of a tensor may have diverse numerical ranges (e.g., outliers in LLM activations). Group-wise quantization adapts to these local variations, reducing quantization error compared to per-tensor methods.

During inference, since the calculation of the softmax function needs high-precision data, we need to dequantize the attention weights matrix after getting it through matrix multiplication. Both the dequantization and matrix multiplication operators require loading matrices from the GPU global memory into the GPU high-speed SRAM, then writing results back to global memory after calculation. Such two operators will cause two read and write operations. To further accelerate the inference process, we apply operator fusion, which fuses the operators of dequantization and matrix multiplication and designs a new operator. The idea of operator fusion is derived from AutoTVM \cite{chen2018tvm}. By fusing these two operators into one, we reduce memory transactions to a single read and write operation, while keeping all intermediate computations within the GPU's high-speed SRAM hierarchy. During the practical inference, we directly input the quantized key matrix and the original precision Q matrix into the new operator, which returns a high-precision floating-point attention weights result for the following softmax function. After the softmax function, we again input the quantized value matrix and the original precision attention weights matrix into the new operator and achieve the high precision attention output. By fusing different operators together, we avoid costly intermediate memory read and write of dequantized matrices, significantly reduce global memory bandwidth pressure, eliminate kernel launch overhead between operations, and enable hardware-friendly computation patterns that maximize cache utilization and inference latency reduction.

\subsubsection{Limitation}

It should be noted that our reordering method still has one limitation: reordering will affect the output result in the prefill stage computation. This is because, after the Q and K matrices are multiplied, the attention weights matrix is produced in the prefill stage, and a mask matrix needs to be added to ensure causal relationships in token generation. The presence of this mask matrix causes the reordered attention weights matrix to yield different results when fed into the softmax function. In this case, the softmax output no longer equals the result obtained by first applying softmax to the original attention weights matrix and then performing reordering. In the decoding stage, reordering the KV cache does not affect the output results since no mask matrix is applied in this stage.

\begin{table}[t]
\centering
\caption{Performance comparison of various models on multi-choice video QA benchmarks. Some of the results are cited from TS-LLaVA \cite{qu2024ts}.}
\label{tab: multivideo_qa}
\begin{tabular}{l|ccccc}
\toprule
Model & LLM Size & VisionEncoder & NExT-QA & EgoSchema & IntentQA  \\
\midrule
Video-LLaVA & 7B & ViT-L & 60.5 & 37.0 & - \\
Video-LLaMA2 & 7B & CLIP-L & - & 51.7 & -   \\
MovieChat+ & 7B & CLIP-G & 54.8 & 56.4 & - \\
Vista-LLaMA & 7B & CLIP-G & 60.7 & - & -  \\
DeepStack-L & 7B & CLIP-L & 61.0 & 38.4 & - \\
$M^3$ & 7B & CLIP-L & 63.1 & 36.8 & 58.8  \\
IG-VLM & 7B & CLIP-L & 63.1 & 35.8 & 60.3 \\
SF-LLaVA & 7B & CLIP-L & 64.2 & 47.2 & 60.1 \\
TS-LLaVA & 7B & CLIP-L & 66.5 & 50.2 & 61.7  \\

LLaVA-OneVision-Qwen2 & 7B & SigLIP & 67.8 & 59.8 & 64.9 \\
\rowcolor{blue!10}
LLaVA-OneVision-Qwen2+WindowQuant & 7B & SigLIP & \textbf{78.9} & \textbf{64.1} & 69.4 \\

Qwen2-VL & 7B & ViT-L & 75.2 & 54.6 &  \textbf{88.6} \\
\rowcolor{blue!10}
Qwen2-VL+WindowQuant & 7B & ViT-L & 74.6 & 53.2 & 87.8 \\

InternVL2 & 8B & InternViT & 68.7 & 48.8 & 67.5 \\
\rowcolor{blue!10}
InternVL2+WindowQuant & 8B & InternViT & 76.1 & 55.8 & 75.9 \\
\bottomrule
\end{tabular}
\end{table}

Therefore, during actual inference, we first compute the results for the first output token in the prefill stage. We then perform reordering and quantization to obtain the reordered, quantized KV cache (storing only this KV cache without recomputing the first output token). This reordered quantized KV cache subsequently participates in all following computations. In the later decoding stage, we can normally perform reordering before computing each output token.


\section{Evaluation}

\begin{table*}[htbp]
\centering
\caption{Performance comparison of different video understanding models on the MVBench dataset. ``LLaVA-OV" means LLaVA-OneVision-Qwen2. Some of the results are cited from TS-LLaVA \cite{qu2024ts}.}
\label{tab: mvbench}
\resizebox{\textwidth}{!}{
\begin{tabular}{l|cccccccccccc}
\toprule
\multirow{2}{*}{Method} & \multirow{2}{*}{\makecell{LLM\\Size}} & \multirow{2}{*}{\makecell{Vision\\Encoder}}  & \multirow{2}{*}{AA} & \multirow{2}{*}{AC} & \multirow{2}{*}{AL} & \multirow{2}{*}{AP} & \multirow{2}{*}{AS} & \multirow{2}{*}{CO} & \multirow{2}{*}{CI} & \multirow{2}{*}{EN} & \multirow{2}{*}{ER} & \multirow{2}{*}{UA} \\
\\
\midrule
Video-ChatGPT  & 7B & CLIP-L & 62.0 & 30.5 & 20.0 & 26.0 & 23.5 & 33.0 & 35.5 & 29.5 & 26.0 & 32.7 \\

Video-LLaMA  & 7B & CLIP-G & 51.0 & 34.0 & 22.5 & 25.5 & 27.5 & 40.0 & 37.0 & 30.0 & 29.0 & 34.1 \\

VideoChat  & 7B & CLIP-G & 56.0 & 35.0 & 27.0 & 26.5 & 33.5 & 41.0 & 36.0 & 32.5 & 33.5 & 35.5 \\

VideoChat2  & 7B & UMT-L & 83.5 & 39.0 & 23.0 & 47.5 & 66.0 & 36.5 & 65.5 & 35.0 & \textbf{49.5} & 51.1 \\

ST-LLM  & 7B & ViT-G & 84.0 & 36.5 & 31.0 & 53.5 & 66.0 & 46.5 & 58.5 & 34.5 & 44.0 & 54.9 \\

PLLaVA & 7B & CLIP-L & 55.5 & 39.5 & 26.0 & 49.0 & 58.0 & 53.5 & 51.0 & 30.5 & 41.0 & 46.6 \\

PLLaVA & 34B & CLIP-L & 82.0 & 40.5 & \textbf{49.5} & 53.0 & 67.5 & \textbf{66.5} & 59.0 & 39.5 & 47.0 & 58.1 \\

GPT-4V & GPT4 & Unknown & 72.0 & 39.0 & 40.5 & 63.5 & 55.5 & 52.0 & 11.0 & 31.0 & 46.5 & 43.5 \\

TS-LLaVA & 7B & CLIP-L & 58.5 & 41.0 & 27.0 & 53.5 & 54.0 & 53.0 & 32.0 & 32.5 & 36.0 & 45.5 \\

TS-LLaVA & 34B & CLIP-L & 73.5 & \textbf{46.5} & 39.0 & 47.5 & 61.0 & 66.5 & 39.0 & \textbf{42.5} & 46.5 & 52.6 \\

LLaVA-OV & 7B & SigLIP & 72.0 & 36.0 & 32.0 & 40.0 & 47.9 & 42.0 & 41.5 & 29.5 & 31.5 & 60.5  \\
\rowcolor{blue!10}
LLaVA-OV+WindowQuant & 7B & SigLIP & 78.5 & 38.0 & 37.0 & 69.5 & 66.5 & 57.5 & 41.0 & 31.0 & 39.5 & \textbf{81.5} \\ 
Qwen2-VL & 7B & ViT-L & \textbf{81.5} & 43.0 & 35.0 & 42.0 & 55.3 & 46.5 & 59.5 & 36.0 & 33.0 & 73.5\\
\rowcolor{blue!10}
Qwen2-VL+WindowQuant & 7B & ViT-L & 76.0 & 42.5 & 35.5 & 64 & 67.0 & 58.0 & 48.5 & 37.5 & 40.5 & 74 \\
InternVL2 & 8B & InternViT & 73.5 & 40.5 & 38.0 & 63.0 & 54.3 & 51.0 & 56.0 & 33.5 & 37.5 & 58.5\\
\rowcolor{blue!10}
InternVL2+WindowQuant & 8B & InternViT & 77.0 & 35.5 & 34.5 & \textbf{70.0} & \textbf{69.1} & 46.5 & \textbf{72.0} & 33.5 & 36.5 & 62.5\\

\midrule

\multirow{2}{*}{Method} & \multirow{2}{*}{\makecell{LLM\\Size}} & \multirow{2}{*}{\makecell{Vision\\Encoder}} & \multirow{2}{*}{FA} & \multirow{2}{*}{MA} & \multirow{2}{*}{MC} & \multirow{2}{*}{MD} & \multirow{2}{*}{OE} & \multirow{2}{*}{OI} & \multirow{2}{*}{OS} & \multirow{2}{*}{ST} & \multirow{2}{*}{SC} & \multirow{2}{*}{Avg.} \\
\\
\midrule
Video-ChatGPT  & 7B & CLIP-L & 22.5 & 39.5 & 25.5 & 23.0 & 54.0 & 28.0 & 40.0 & 31.0 & 48.5 & 32.7 \\

Video-LLaMA  & 7B & CLIP-G & 29.0 & 32.5 & 22.5 & 22.5 & 48.0 & 40.5 & 38.0 & 43.0 & 45.5 & 34.1 \\

VideoChat  & 7B & CLIP-G & 33.5 & 42.5 & 20.5 & 31.5 & 53.0 & 40.5 & 30.0 & 48.5 & 46.0 & 35.5 \\

VideoChat2  & 7B & UMT-L & 49.5 & 58.5 & 42.0 & 23.0 & 58.0 & 71.5 & \textbf{42.5} & 88.5 & 44.0 & 51.1 \\

ST-LLM  & 7B & ViT-G & 44.0 & 78.5 & 56.5 & 42.5 & 80.5 & 73.5 & 38.5 & 86.5 & 43.0 & 54.9 \\

PLLaVA & 7B & CLIP-L & 41.0 & 52.0 & 42.0 & 23.5 & 56.0 & 61.0 & 36.0 & 82.0 & 45.0 & 46.6 \\

PLLaVA & 34B & CLIP-L & 47.0 & 70.0 & 43.0 & 37.5 & 68.5 & 67.5 & 36.5 & 91.0 & \textbf{51.5} & 58.1 \\

GPT-4V & GPT4 & Unknown & 46.5 & 22.5 & 12.0 & 12.0 & 18.5 & 59.0 & 29.5 & 83.5 & 45.0 & 43.5 \\

TS-LLaVA & 7B & CLIP-L & 36.0 & 49.0 & 30.0 & 23.5 & 58.5 & 59.5 & 32.0 & 85.0 & 43.5 & 45.5 \\

TS-LLaVA & 34B & CLIP-L & 46.5 & 51.5 & 30.0 & 24.0 & 52.5 & 65.5 & 37.5 & 89.5 & 45.5 & 52.6 \\

LLaVA-OV & 7B & SigLIP & 42.5 & 60.0 & 38.5 & 28.5 & 59.1 & 54.0 & 36.0 & 69.5 & 42.5 & 45.5\\
\rowcolor{blue!10}
LLaVA-OV+WindowQuant & 7B & SigLIP & 44.5 & 63.5 & 43.0 & 28.5 & 52.5 & \textbf{77.0} & 36.0 & 90.5 & 48.5 & 53.9 \\ 
Qwen2-VL & 7B & ViT-L & \textbf{52.0} & 81.0 & 63.0 & 35.0 & 79.8 & 53.5 & 37.0 & 81.0 & 45.5 & 54.4 \\
\rowcolor{blue!10}
Qwen2-VL+WindowQuant & 7B & ViT-L & 45.5 & 69.0 & 48.0 & 33.5 & 64.1 & 68.0 & 34.0 & \textbf{93.0} & 44.0 & 54.9\\
InternVL2 & 8B & InternViT & 40.0 & 78.0 & \textbf{57.0} & 43.0 & 76.3 & 65.0 & 32.0 & 78.0 & 42.0 & 53.5\\
\rowcolor{blue!10}
InternVL2+WindowQuant & 8B & InternViT & 43.0 & \textbf{93.0} & 56.0 & \textbf{48.0} & \textbf{93.4} & 70.0 & 36.0 & 84.0 & 44.5 & \textbf{58.2} \\

\bottomrule
\end{tabular}
}
\end{table*}

\begin{table}[h]
\centering
\caption{The performance comparison of WindowQuant with other VLM quantization methods. Some of the results are cited from AKVQ \cite{su2025akvq}.}
\resizebox{\textwidth}{!}{
\begin{tabular}{c|c|cccccccccccc}
\toprule
\textbf{Model} & \textbf{Method} & \textbf{OE} & \textbf{OI} & \textbf{MA} & \textbf{EN} & \textbf{SC} & \textbf{ST} & \textbf{WQA} & \textbf{TQA} & \textbf{MQA} & \textbf{SQA} & \textbf{DQA} & \textbf{Avg} \\
\midrule
\multirow{7}{*}{LLaVA-v1.5-7B} & FP16 & 53.0 & 46.5 & 47.5 & 33.5 & 36.5 & 73.0 & 59.0 & 47.0 & 68.5 & 43.0 & 45.5 & 50.3 \\
\cmidrule{2-14}
& RTN (INT4) & 46.0 & 43.5 & 47.5 & \textbf{28.5} & \textbf{35.5} & 58.5 & 43.0 & \textbf{43.5} & 62.5 & 41.5 & 37.5 & 44.4 \\
& RTN (INT2) & 15.5 & 12.0 & 12.0 & 13.5 & 8.0 & 7.5 & 7.0 & 16.0 & 13.5 & 9.5 & 5.5 & 10.9 \\

& SmoothQuant & 37.5 & 24.0 & 24.5 & 22.0 & 31.5 & 24.5 & 24.0 & 25.5 & 26.5 & 25.5 & 21.0 & 26.0 \\
& KIVI & 18.0 & 16.0 & 24.0 & 20.0 & 21.5 & 16.0 & 13.0 & 18.5 & 27.5 & 19.0 & 18.0 & 19.3 \\
& SKVQ & 35.0 & 39.0 & 14.0 & 20.0 & 40.5 & 48.5 & 24.5 & 20.5 & 54.5 & 27.0 & 13.5 & 30.6 \\
\rowcolor{blue!10}
\cellcolor{white} & WindowQuant  & \textbf{50.5} & \textbf{48.5} & \textbf{49.0} & 27.5 & 33.5 & \textbf{65.5} & \textbf{58.5} & 43.0 & \textbf{67.0} & \textbf{46.5} & \textbf{44.0} & \textbf{48.5} \\

\midrule

\multirow{7}{*}{LLaVA-v1.5-13B} & FP16 & 46.0 & 45.0 & 50.0 & 26.5 & 37.0 & 70.5 & 64.0 & 55.0 & 74.5 & 51.0 & 46.0 & 51.4 \\
\cmidrule{2-14}
& RTN (INT4) & \textbf{49.0} & 42.0 & 47.5 & \textbf{28.5} & 35.0 & 65.5 & 64.0 & \textbf{51.5} & \textbf{71.0} & 45.0 & 47.0 & 49.6 \\
& RTN (INT2) & 27.0 & 12.0 & 19.5 & 11.5 & 20.0 & 8.5 & 6.5 & 15.0 & 19.5 & 14.5 & 13.0 & 15.2 \\

& SmoothQuant & 36.0 & 26.5 & 23.0 & 21.0 & 31.0 & 25.0 & 24.0 & 27.0 & 26.5 & 25.5 & 24.5 & 26.4 \\
& KIVI & 28.5 & 25.5 & 38.0 & 26.5 & 22.0 & 41.0 & 34.0 & 34.0 & 45.5 & 29.5 & 35.5 & 32.7 \\
& SKVQ & 49.0 & 41.5 & 40.0 & 25.5 & \textbf{40.5} & 45.5 & 45.5 & 30.0 & 48.5 & 31.5 & 30.5 & 38.9 \\
\rowcolor{blue!10}
\cellcolor{white} & WindowQuant & 47.5 & \textbf{47.0}  & \textbf{50.0} & 26.5 & 36.5 & \textbf{75.0} & \textbf{65.0}  & 49.5 & 70.5 & \textbf{53.0} & \textbf{47.0} & \textbf{51.6} \\

\midrule

\multirow{7}{*}{LLaVA-v1.6-vicuna-7B} & FP16 & 29.0 & 18.0 & 18.5 & 28.0 & 20.0 & 63.0 & 50.5 & 41.5 & 38.5 & 42.0 & 41.5 & 35.5 \\
\cmidrule{2-14}
& RTN (INT4) & 33.5 & 8.5 & \textbf{30.0} & 18.5 & 17.0 & \textbf{61.5} & 27.0 & 33.0 & 22.5 & 27.5 & 30.5 & 28.1 \\
& RTN (INT2) & 8.0 & 4.0 & 3.0 & 4.5 & 2.5 & 4.0 & 6.5 & 10.5 & 3.5 & 6.5 & 5.0 & 5.3 \\

& SmoothQuant & 35.0 & 13.5 & 23.5 & 21.5 & 17.5 & 26.5 & 25.5 & 24.5 & 28.5 & 24.5 & 24.5 & 24.1 \\
& KIVI & 17.5 & 15.5 & 27.0 & 17.5 & 14.5 & 35.0 & 18.5 & 15.5 & 21.0 & 12.5 & 14.5 & 19.0 \\
& SKVQ & 34.0 & \textbf{26.5} & 23.5 & 11.0 & \textbf{32.5} & 32.0 & 31.0 & 11.5 & \textbf{33.0} & 23.0 & 23.0 & 25.5 \\
\rowcolor{blue!10}
\cellcolor{white} & WindowQuant  & \textbf{43.0} & 20.0 & 19.0 & \textbf{28.0} & 14.5 & 56.0 & \textbf{41.5} & \textbf{41.5} & 30.0 & \textbf{39.0} & \textbf{43.5} & \textbf{34.2} \\

\midrule

\multirow{7}{*}{LLaVA-v1.6-mistral-7B} & FP16 & 44.5 & 46.0 & 44.0 & 34.5 & 38.0 & 71.0 & 55.5 & 49.5 & 66.0 & 44.0 & 38.0 & 48.3 \\
\cmidrule{2-14}
& RTN (INT4) & 35.5 & 24.0 & 26.0 & 28.0 & 20.0 & 63.0 & 50.5 & 41.5 & 38.5 & 42.0 & 41.5 & 37.3 \\
& RTN (INT2) & 29.0 & 18.0 & 18.5 & 20.0 & 18.0 & 11.0 & 13.5 & 18.0 & 25.5 & 17.5 & 22.5 & 19.2 \\

& SmoothQuant & 38.5 & 26.0 & 26.0 & 24.0 & 33.0 & 45.0 & 29.0 & 45.5 & 27.0 & 34.0 & 33.5 & 32.9 \\
& KIVI & 45.5 & 42.5 & 47.5 & 24.0 & 36.0 & 53.0 & 47.0 & 49.5 & 62.5 & 39.0 & 37.5 & 44.0 \\
& SKVQ & \textbf{49.0} & 36.5 & 39.5 & 27.5 & 34.5 & 56.0 & 46.5 & 47.5 & 52.5 & 37.0 & 33.5 & 41.8 \\
\rowcolor{blue!10}
\cellcolor{white} & WindowQuant & 43.0 & \textbf{52.0} & \textbf{55.5} & \textbf{34.0} & \textbf{38.5} & \textbf{70.0} & \textbf{58.5} & \textbf{51.0} & \textbf{66.0} & \textbf{44.5} & \textbf{45.5} & \textbf{50.8}  \\
\bottomrule
\end{tabular}}

\label{tab: quantization}
\end{table}

\subsection{Experiment Setup}

\textbf{Benchmarks.} We begin our evaluation by conducting a comprehensive comparison of WindowQuant's question-answering (QA) capabilities against several SOTA VLMs across multiple standardized QA benchmarks. Our assessment encompasses three widely-recognized multi-choice QA datasets: NExT-QA, which focuses on temporal and causal reasoning in video understanding; EgoSchema, designed to evaluate long-form video comprehension from an egocentric perspective; and IntentQA, which tests models' ability to infer human intentions and motivations from visual content. This multi-benchmark evaluation framework allows us to thoroughly assess WindowQuant's performance across diverse question-answering scenarios, ranging from temporal reasoning and causal inference to intent recognition and long-form video understanding. 

We further use a multiple-task video understanding dataset called MVBench to compare the performance between WindowQuant and SOTA VLMs comprehensively and thoroughly. The benchmark is MVBench, which is a comprehensive video understanding benchmark. We select 19 challenging video tasks from MVBench, containing nine types of reasoning abilities in video understanding. The details of the 19 video tasks in MVBench are as follows: AA (Action Antonym),  AC (Action Count),  AL (Action Localization),  AP (Action Prediction),  AS (Action Sequence),  CO (Character Order), CI (Counterfactual Inference), EN (Egocentric Navigation), ER (Episodic Reasoning), FA (Fine-grained Action), MA (Moving Attribute), MC (Moving Count), MD (Moving Direction), OE (Object Existence), OI (Object Interaction), OS (Object Shuffle), ST (Scene Transition), SC (State Change), UA (Unexpected Action). It is worth noting that the videos in MVBench are generally short, and the evaluation on the MVBench dataset also reflects the capability of WindowQuant on short-video tasks.

We also use a comprehensive benchmark, MileBench \cite{song2024milebench}, to compare the performance of WindowQuant with other quantization methods. We select 12 tasks from MileBench, including Object Existence (OE), Object Interaction (OI),
Moving Attribute (MA), Egocentric Navigation (EN), State
Change (SC), Scene Transition (ST), Space Understanding(SU), Webpage QA (WQA), Textbook QA (TQA), multimodal
QA (MQA), Slide VQA (SQA), and Document QA (DQA).

\textbf{Baselines.}
We compare the performance of WindowQuant with several SOTA VLM models in recent years, including Video-LLaVA \cite{lin2023video}, MovieChat+ \cite{song2025moviechat+}, Vista-LLaVA \cite{ma2024vista}, DeepStack-L \cite{meng2024deepstack}, $M^3$ \cite{cai2024matryoshka}, IG-VLM \cite{kim2024image}, SF-LLaVA \cite{xu2024slowfast}, TS-LLaVA \cite{qu2024ts}, LLaVA-OneVision-Qwen2 \cite{li2024llava}, Qwen2-VL \cite{wang2024qwen2}, InternVL2 \cite{chen2024far}, Video-ChatGPT \cite{maaz2023video}, Video-LLaMA \cite{zhang2023video}, VideoLLaMA2 \cite{cheng2024videollama}, VideoChat \cite{li2024mvbench}, ST-LLM \cite{liu2024st}, PLLaVA \cite{xu2024pllava}, GPT-4V \cite{openai2023gpt4v}. The size of most of the LLMs in these VLMs is 7B, while a few of them are 34B.
We also compare the performance of WindowQuant with basic uniform RTN  quantization method and some SOTA VLM quantization methods, including SmoothQuant \cite{xiao2023smoothquant}, KIVI \cite{liu2024kivi}, SKVQ \cite{duanmu2024skvq}.

\textbf{Implementation.} 
We evaluate WindowQuant on three famous VLM models on the multi-choice QA datasets: LlaVA-OneVision-Qwen2 \cite{li2024llava}, Qwen2-VL \cite{wang2024qwen2}, and InternVL2 \cite{chen2024far}. Then we implement WindowQuant on four VLM models on the MVBench dataset: LLaVA-v1.5-7B, LLaVA-v1.5-13B, LLaVA-v1.6-vicuna-7B, LLaVA-v1.6-mistral-7B \cite{liu2023visual}. The original full-precision versions of these models (i.e., the FP16 models) only adopt the standard FlashAttention optimization and do not employ any additional optimization techniques.
We conduct all the experiments on an NVIDIA A800 GPU containing 80GB of GPU memory, with a 25-core AMD EPYC 7T83 CPU and 100GB of memory. Unless otherwise specified, in our experiments we generally set the threshold parameter to 2, the window size to 32, the batch size to 1, and extract 100 video frames.

\begin{figure*}[t]
    \centering
    \subfigure[]{
    \label{fig: memory}
    \begin{minipage}[b]{0.45\linewidth}
        \centering
        \includegraphics[width=1\linewidth]{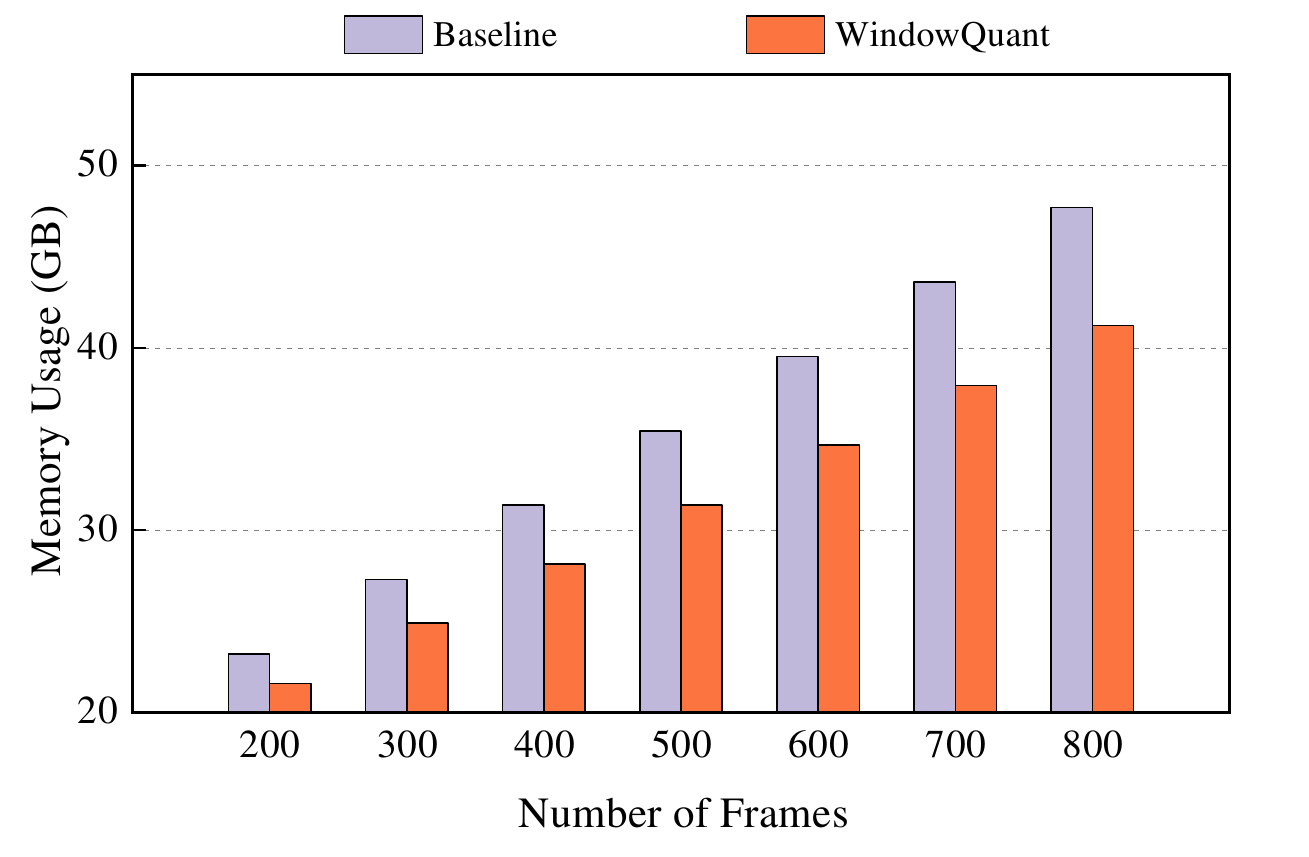}
    \end{minipage}}
    \hspace{0.05\linewidth}
    \subfigure[]{
    \label{fig: throughput}
    \begin{minipage}[b]{0.45\linewidth}
        \centering
        \includegraphics[width=1\linewidth]{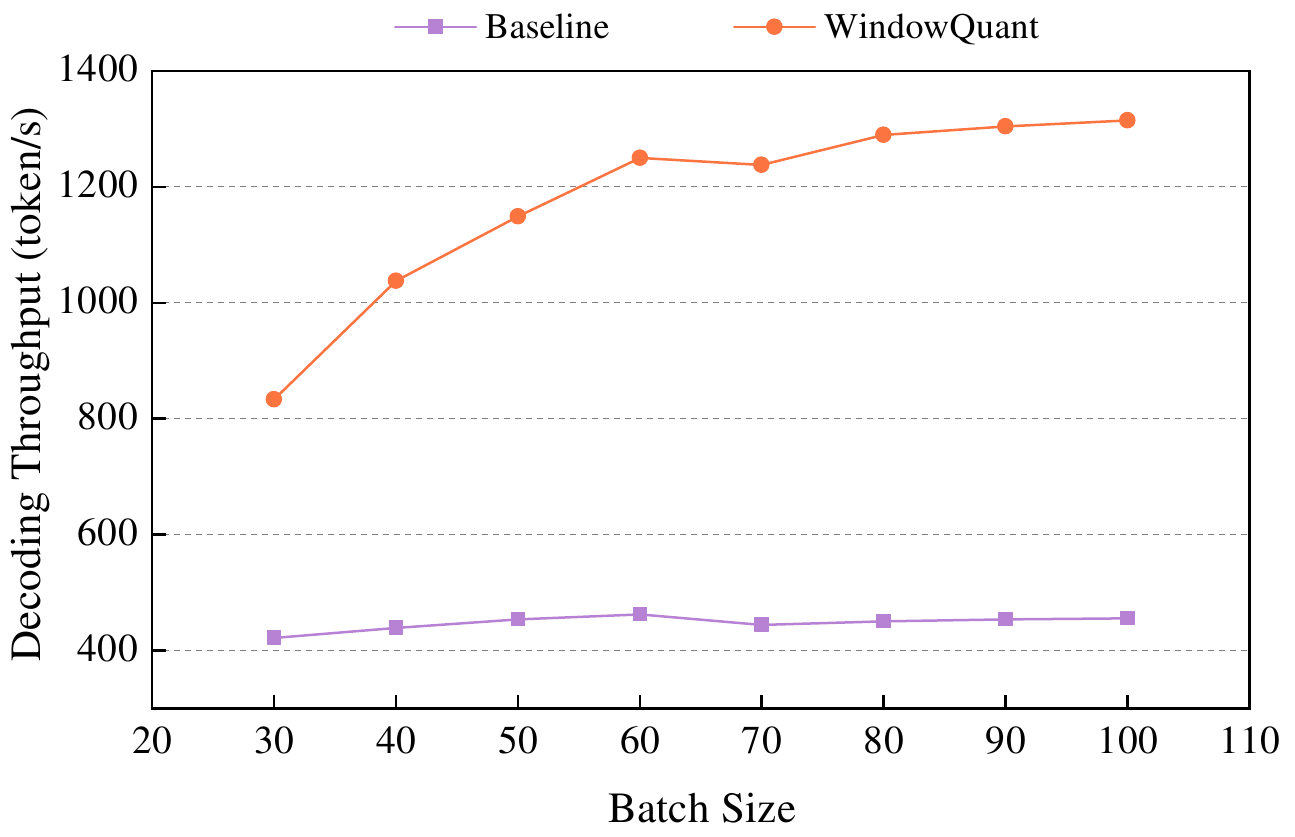}
    \end{minipage}}
    \Description{Diagram showing the efficiency test.}
    \caption{(a): GPU memory usage comparison of WindowQuant and FP16 model under different number of frames. (b): Decoding throughput comparison of WindowQuant and FP16 model under different batch sizes.}
\end{figure*}

\subsection{Main Results}

\begin{table}[]
    \centering
    \caption{GPU memory usage comparison of WindowQuant and other quantization methods.}
    \begin{tabular}{c|ccccccc}
    \toprule
    Method     &  FP16 & RTN (INT4) & RTN (INT2) & SmoothQuant & KIVI & SKVQ & WindowQuant \\
    \midrule
    Average Bit-width & 16 & 4 & 2 & 2 & 2 & 2 & 3.17 \\
    GPU Memory Usage (GB) & 19.13 & 18.40 & 18.17 & 18.17 & 18.17 & 18.17    & 18.33 \\
    \bottomrule
    \end{tabular}
    
    \label{tab: quantization memory comparison}
\end{table}

\begin{table}[]
    \centering
    \caption{The latency of different processes during inference of FP16 model and WindowQuant.}
    \resizebox{\textwidth}{!}{\begin{tabular}{c|cccccc}
    \toprule
    Method & Input Preparation & Feature Encoding & Similarity Calculation & Prefill Stage & Decoding Stage & Total \\
    \midrule
        FP16 & 45 ms & 920 ms & 0 ms & 3098 ms & 2451 ms & 6514 ms \\
        WindowQuant & 24 ms & 920 ms & 31 ms & 3099 ms & 1640 ms & 5714 ms \\
        \bottomrule
    \end{tabular}}
    
    \label{tab:latency}
\end{table}

\begin{figure}[t]
  \centering
  \begin{minipage}[t]{0.6\textwidth}
  \vspace{0pt}
    \centering
    \includegraphics[width=\linewidth]{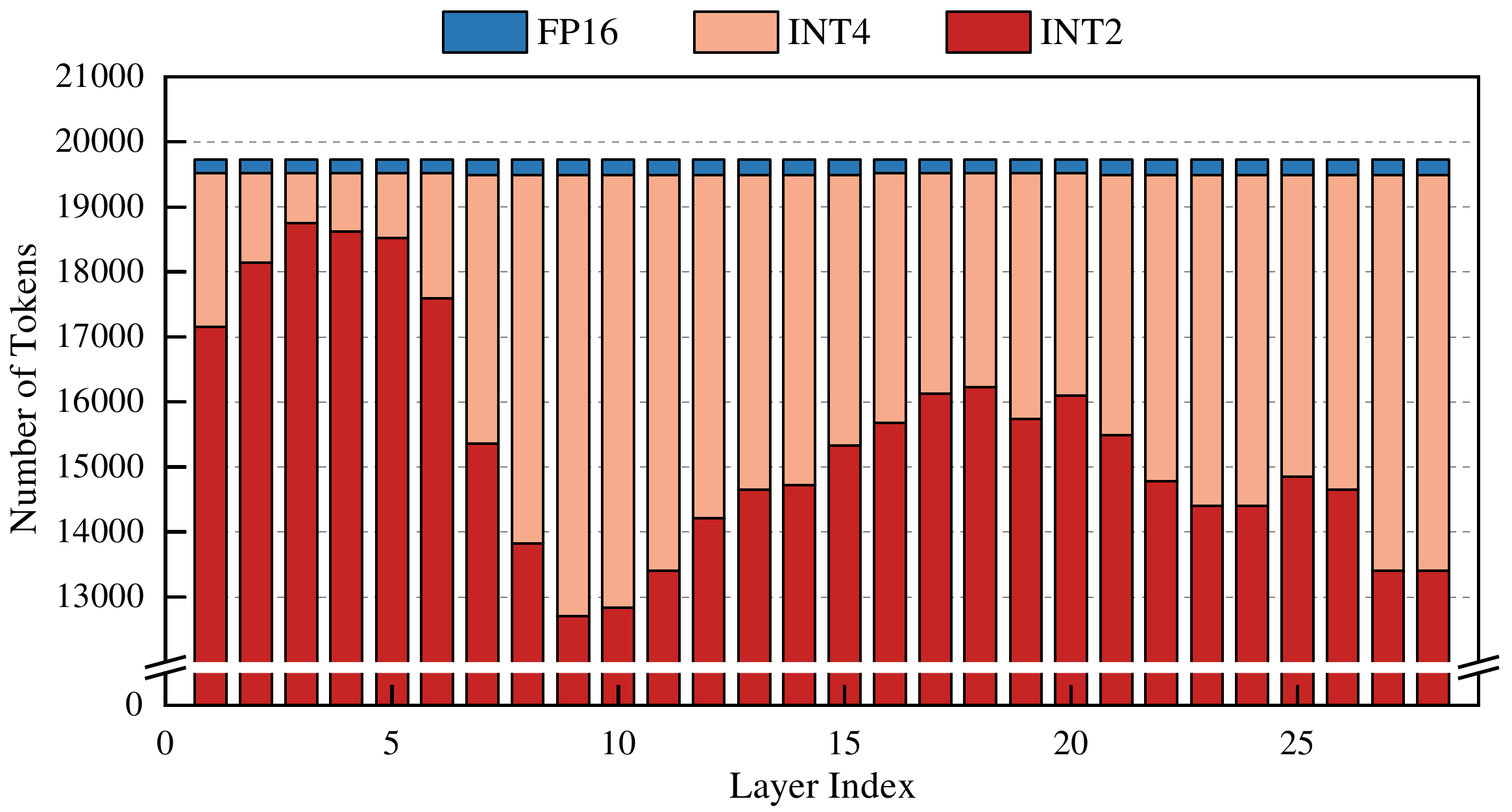}
    \caption{The number of tokens with different bit-widths in each layer after applying WindowQuant with 100 input frames.}
    \label{fig: proportion}
  \end{minipage}
  \hspace{0.03\linewidth}
  \begin{minipage}[t]{0.33\textwidth}
  \vspace{0pt}
    \centering
    \captionof{table}{The memory usage of different data with 100 input frames.}
    \label{tab: memory proportion}
    \resizebox{1\textwidth}{!}{
    \begin{tabular}{ccc}
      \toprule
      Data  & FP16 & WindowQuant \\
      \midrule
      Model weights & 15.01 GB & 15.01 GB \\ 

      \midrule
      Encoded features & 0.57 GB & 0.57 GB \\
      \midrule
      \makecell{Intermediate tensors \\ in multimodal encoding} & 1.7 GB & 1.7 GB \\
      \midrule
      \makecell{Intermediate tensors \\ in FFN} & 2.22 GB & 2.22 GB \\
      \midrule
      \makecell{Intermediate tensors \\ in LayerNorm} & 0.14 GB & 0.14 GB \\
      \midrule
      KV cache & 1.05 GB & 0.19 GB \\
      \midrule
      Quantization metadata & 0 GB & 0.06 GB \\
      \midrule
      Output hidden states & 0.14 GB & 0.14 GB \\
      \midrule
      Peak memory usage & 19.13 GB & 18.33 GB \\
      \bottomrule
    \end{tabular}}
  \end{minipage}
  \vspace{-5mm}
\end{figure}

    
    

\begin{figure*}[t]
    \centering
    \subfigure[]{
    \label{fig: window_size_latency}
    \begin{minipage}[b]{0.45\linewidth}
        \centering
        \includegraphics[width=1\linewidth]{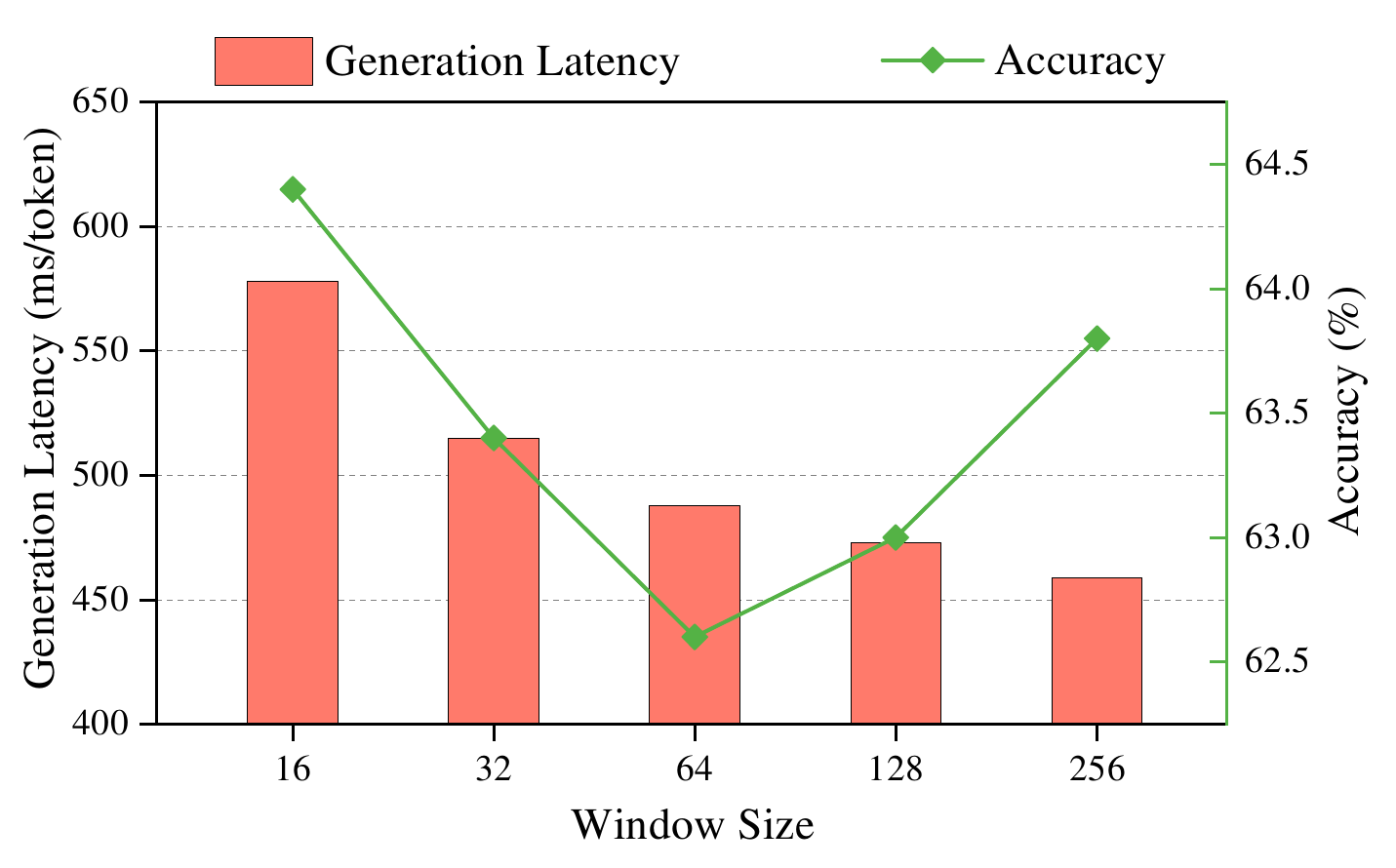}
    \end{minipage}}
    \hspace{0.05\linewidth}
    \subfigure[]{
    \label{fig: window_size_proportion}
    \begin{minipage}[b]{0.45\linewidth}
        \centering
        \includegraphics[width=1\linewidth]{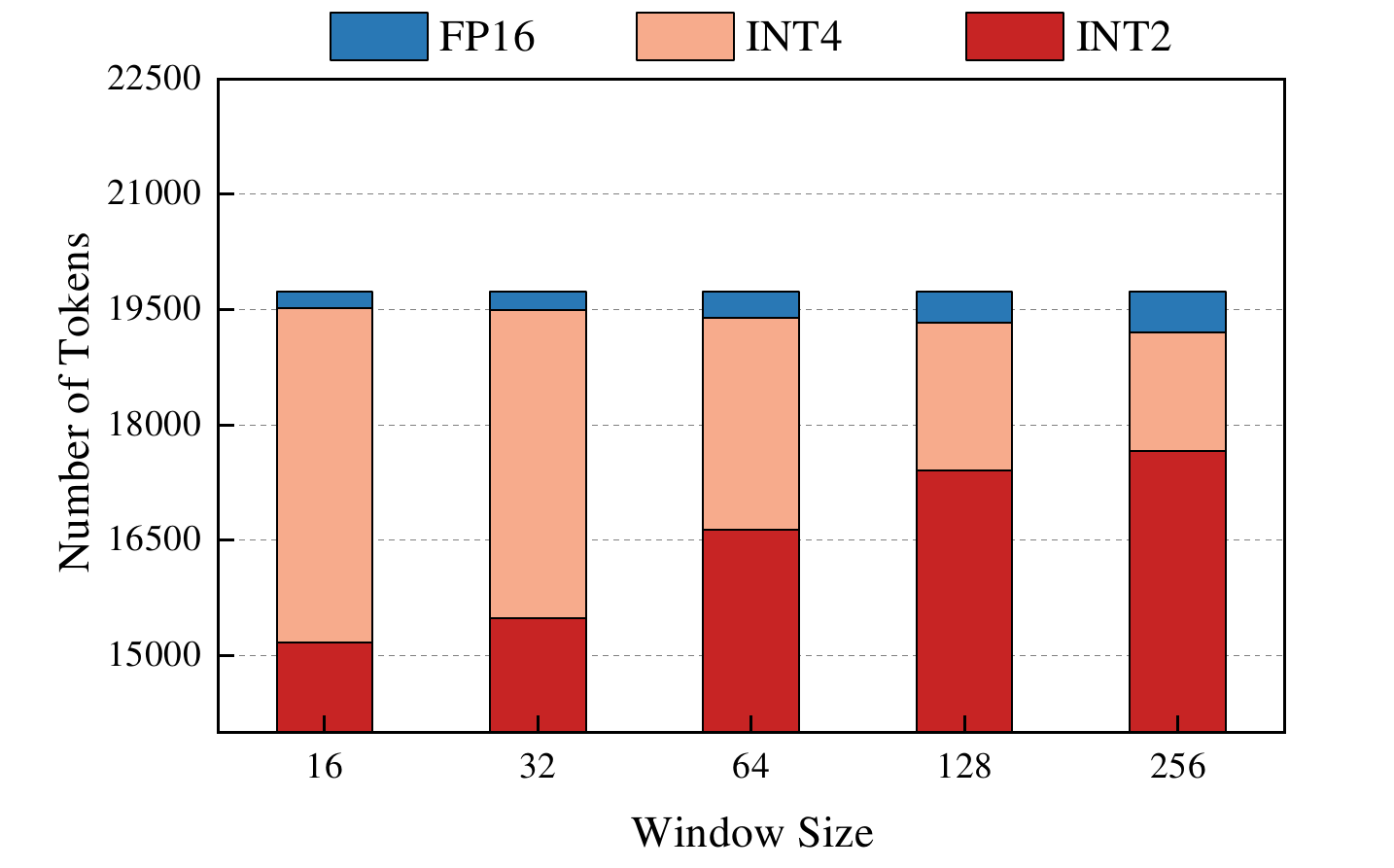}
    \end{minipage}}
    \caption{(a): The generation latency of WindowQuant under different window sizes. (b): The number of tokens with different bit-widths of WindowQuant under different window sizes.}
    \Description{Diagram showing the window size comparison.}
    \vspace{-5mm}
    
\end{figure*}

First, we compare the performance of WindowQuant with the baseline VLM models over the three different QA datasets. For the IntentVL2 model, we extract 25 video frames in our experiments because its feature extraction code consumes a large amount of GPU memory, and extracting more frames would lead to OOM (out-of-memory) errors. The results are shown in Table \ref{tab: multivideo_qa}. Our WindowQuant method demonstrates significant performance improvements across multiple benchmark evaluations. On the NExT-QA and EgoSchema datasets, the models with the WindowQuant method achieve the best performance. Specifically, WindowQuant serves as a quantization technique that achieves consistent enhancements when applied to different base models:
For the LLaVA-OneVision-Qwen2 model, WindowQuant substantially improves NExT-QA performance from 67.8 to 78.9, EgoSchema from 59.8 to 64.1, and IntentQA from 64.9 to 69.4. When applied to the Qwen2-VL model, WindowQuant shows a slight adjustment in NExT-QA from 75.2 to 74.6, while maintaining competitive performance on IntentQA (88.6 to 87.8). Notably, the InternVL2 model with WindowQuant achieves remarkable improvements: NExT-QA increases from 68.7 to 76.1, EgoSchema from 48.8 to 55.8, and IntentQA from 67.5 to 75.9. 

Then, we conduct the performance experiments on the multi-task video understanding dataset MVBench. As illustrated in Table \ref{tab: mvbench}, across all sub-tasks, the WindowQuant method universally achieves excellent performance. The highest average accuracy is obtained on the InternVL2 model with WindowQuant. After applying WindowQuant, the LLaVA-OneVision-Qwen2 model achieves notable average performance improvement (59.8 to 64.1), the Qwen2-VL model maintains a competitive average accuracy (54.4 to 54.9), and InternVL2 also delivers an impressive average performance improvement (53.5 to 58.2).
These results indicate that WindowQuant, as a quantization optimization strategy, can effectively enhance video understanding task performance with even lower precision data. This is because we use window-level mixed-precision quantization to successfully maintain the precision of those important video parts. More importantly, since the model's input contains some redundant information and even noise, our quantization method can appropriately mitigate such information, thereby enabling the model to achieve better performance than the original full precision. Notably, performance degradation is observed on some tasks (e.g. MA, MC, OE), while improvements are achieved on others. This behavior arises from the task-dependent sensitivity to the precision of visual token representations. For datasets that require fine-grained, local, or transient visual information, the group-wise quantization adopted by WindowQuant enforces shared quantization parameters within each window, which reduces the numerical discriminability among tokens. This representation compression amplifies quantization errors in such tasks, leading to degraded model performance. In contrast, for tasks whose inputs span multiple frames and primarily rely on global or redundant temporal cues rather than precise single-token values, the impact of quantization errors is less pronounced. Moreover, WindowQuant effectively suppresses noise and emphasizes stable patterns in these scenarios, resulting in overall performance improvements.

We also compare the performance of WindowQuant with other VLM quantization methods. As illustrated in Table \ref{tab: quantization}, WindowQuant achieves the highest average accuracy over these quantization methods on all four models. On certain models (such as LLaVA-v1.5-13B and LLaVA-1.6-mistral-7B), WindowQuant similarly demonstrates better performance than full precision.

\begin{figure*}
    \centering
    \subfigure[]{
    \label{fig: thre_para}
    \begin{minipage}[b]{0.45\linewidth}
        \centering
        \includegraphics[width=1\linewidth]{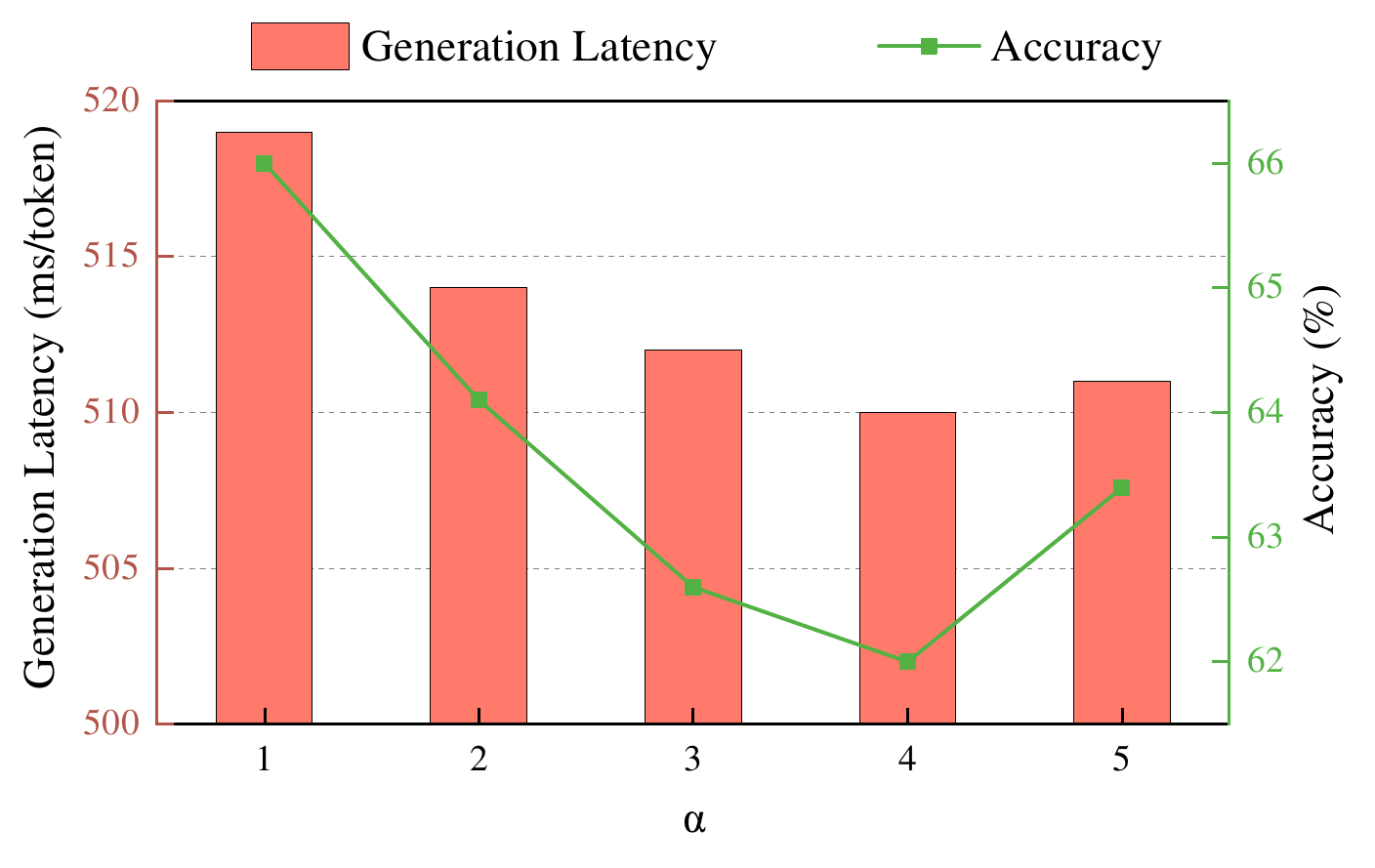}
    \end{minipage}}
    \hspace{0.05\linewidth}
    \subfigure[]{
    \label{fig: thre_para_proportion}
    \begin{minipage}[b]{0.45\linewidth}
        \centering
        \includegraphics[width=1\linewidth]{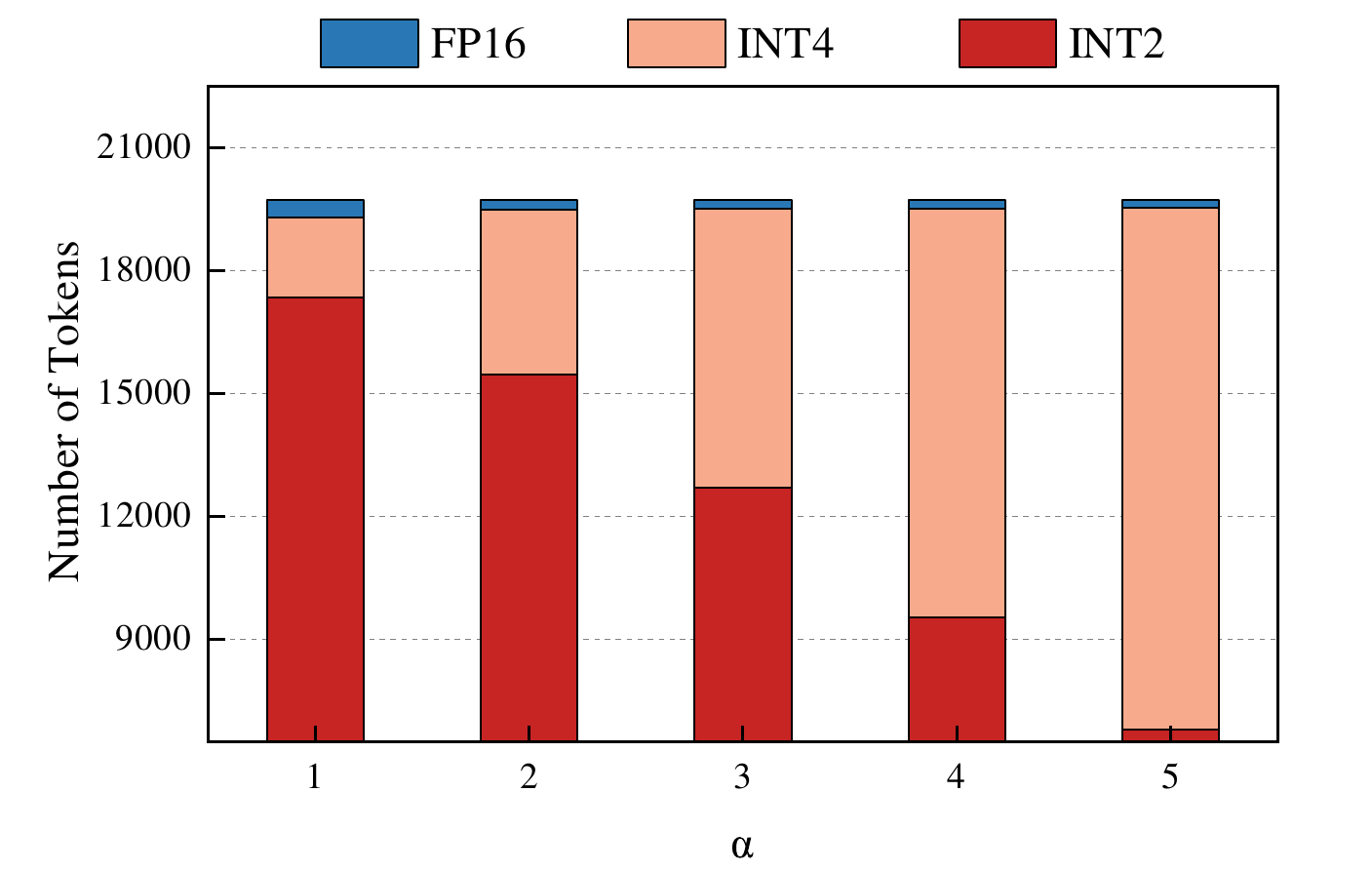}
    \end{minipage}}
    \vspace{-3mm}
    \caption{(a): The generation latency of WindowQuant under different threshold parameter $\alpha$. (b): The number of tokens with different bit-widths of WindowQuant under different threshold parameters $\alpha$.}
    \vspace{-5mm}
    \Description{Diagram showing the threshold parameter comparison.}
\end{figure*}

Furthermore, we compare the GPU memory usage and decoding throughput of WindowQuant with the full precision model on the EgoSchema dataset. As shown in Figure \ref{fig: memory}, WindowQuant can reduce the GPU memory usage by 1.61 GB to 6.48 GB compared to the original FP16 precision model as the number of extracted video frames increases. Notably, the decoding throughput means the throughput during the decoding stage only), which excludes the prefill stage and the preceding visual encoder computation. The reason why we choose decoding throughput as an evaluation metric is that, in practical VLM inference, the decoding stage typically dominates the overall runtime, especially for long output sequences and multi-turn dialogue, where a large number of decoding steps are required and the visual encoding and prefill stage computations can be reused across multiple turns. As the number of input video frames increases, the GPU memory usage reduction effect of WindowQuant becomes increasingly significant. This is because a larger number of input frames leads to a larger KV cache, leaving more room for optimization. We evaluate inputs up to 800 frames, as most videos in our selected datasets contain no more than 800 frames. In real-world applications, the number of input frames can be far larger, in which case WindowQuant is expected to achieve even greater memory savings. 
As for decoding throughput, we extract 5 video frames to avoid OOM error in the experiment. From Figure \ref{fig: throughput} we can see that WindowQuant significantly improves the decoding throughput of the original model. Moreover, the throughput of the original model quickly reaches its upper limit, whereas the decoding throughput of WindowQuant continues to increase with the batch size and does not reach its limit until the batch size exceeds 100 (the results for batch sizes larger than 100 are not plotted due to GPU memory constraints). Compared with the FP16 baseline, our method reduces GPU memory consumption by approximately 15\%, while improving decoding throughput by about 3×. This substantial decoding throughput gain arises not only from the reduced memory footprint, which shortens the time required to load the KV cache from global memory, but also from the more efficient computation enabled by quantized representations (integer arithmetic is more computationally efficient than floating-point arithmetic.). We also compare the memory usage and average bit-width of WindowQuant and other quantization methods (batch size = 1, number of frames = 100). As shown in Table \ref{tab: quantization memory comparison}, the memory usage of WindowQuant is lower than RTN (INT4) but higher than other methods. This is because the average bit-width of WindowQuant is about 3.17, while most methods is 2 bit-width (WindowQuant is a mixed-precision quantization method, while other methods are uniform-precision quantization). However, the memory increase is not significant (only 0.16 GB). Considering the improved accuracy of WindowQuant compared to other quantization methods, we believe this increase in memory usage is acceptable.

To investigate which stages of the inference pipeline are optimized by our method, we further compare the latency breakdown of the FP16 baseline and WindowQuant during inference (batch size = 20, number of frames = 5), as summarized in Table \ref{tab:latency}. In the input preparation stage, WindowQuant achieves a 21 ms reduction in latency compared to FP16, while both methods exhibit identical latency in the feature encoding stage. WindowQuant incurs an additional 31 ms overhead for similarity computation. During the prefill stage, although quantized data in WindowQuant can be loaded and processed more efficiently, the extra quantization operations introduce slight overhead, resulting in a marginal increase of 1 ms in latency. In the decoding stage, WindowQuant reduces latency by 811 ms (in this experiment, we generate 50 tokens for each request; greater reductions can be expected with longer generation lengths). Overall, WindowQuant achieves a total latency reduction of 800 ms.

\subsection{Analysis}
\label{subsec:analysis}

We apply WindowQuant on Llava-Onevision-Qwen2-7B model and conduct experiments on the EgoSchema dataset to analyze the proportion of tokens with different bit-widths in the KV cache in each layer with 100 input frames after applying WindowQuant. As shown in Figure \ref{fig: proportion}, the number of tokens in INT2 precision is the largest, while that in FP16 precision is the smallest. Moreover, the number of FP16 tokens remains relatively stable across layers. This aligns with our expectations, as only a small portion of visual tokens in each layer is related to the text prompt. 

To further study the relationship between token bit-width and memory reduction, we test the main memory usage of different data components during the inference process with 100 input frames. The main memory usage comes from model weights, encoded features, intermediate tensors in visual encoding, intermediate tensors in FFN, intermediate tensors in LayerNorm, KV cache, and quantization metadata. The results are shown in Table \ref{tab: memory proportion}. Note that the intermediate tensors in vision encoding only appear during the multimodal encoding stage, while the other two types of intermediate tensors appear only during the prefill stage. The peak GPU memory usage occurs during the prefill stage, which does not include the intermediate tensors in vision encoding. The difference between FP16 model and WindowQuant lies in KV cache and quantization metadata. WindowQuant can reduces the memory usage by 0.86 GB with 100 input frames, while introduces a slight cost of quantization metadata of 0.06 GB. In total, WindowQuant can reduces the memory usage of FP16 model by 0.8 GB, corresponds with the results in Figure \ref{fig: memory}. Note that these results are obtained with an input of 100 frames. As shown in Figure \ref{fig: proportion}, the average number of INT2 tokens per layer is 12,564, the number of INT4 tokens is 6,956, and only 215 tokens remain in FP16. Accordingly, the average memory reduction per layer is $128*4*2*(12564*\frac{7}{4}+6956*\frac{3}{2})=33199104\text{Byte}\approx 31.66\text{MB}$, resulting in a total memory reduction of $31.66*28=886.48\text{MB}\approx 0.86\text{GB}$ across all layers, which corresponds with the result in Table \ref{tab: memory proportion}. Note that, since the minimum number of video frames in Figure \ref{fig: memory} is 200, while the number of video frames tested here is 100, the memory savings observed here differ from those reported in Figure \ref{fig: memory}.

Nevertheless, the intermediate activations in FFN module remain substantial (These data are no longer existing in the decoding stage, but they will still influence peak memory usage), which limits the overall memory reduction proportion achieved by our method. Optimizing the memory footprint of these intermediate activations will be explored in our future work.

We also evaluate the impact of different window sizes on the model. As shown in Figure \ref{fig: window_size_latency}, the latency of generating tokens gradually decreases as the window size increases. This is because a larger window size allows for faster quantization, since quantization is performed for all tokens within a window at once. Meanwhile, the model’s accuracy first decreases and then increases. To achieve a good balance between latency and accuracy, we choose 32 as the window size value used in our main results.
We further examine the distribution of tokens with different precisions under various window sizes, as shown in Figure \ref{fig: window_size_proportion}. It can be observed that as the window size increases, the numbers of FP16 and INT2 tokens gradually increase, while the number of INT4 tokens gradually decreases. This is because the important video frames within a large window can be surrounded by many unimportant frames, leading to more windows being quantized to INT2. On the other hand, since we fix the precision of the first window as FP16, the number of FP16 tokens will increase slightly as the window size increases. 

\begin{figure}[t]
    \centering
    \includegraphics[width=1\linewidth]{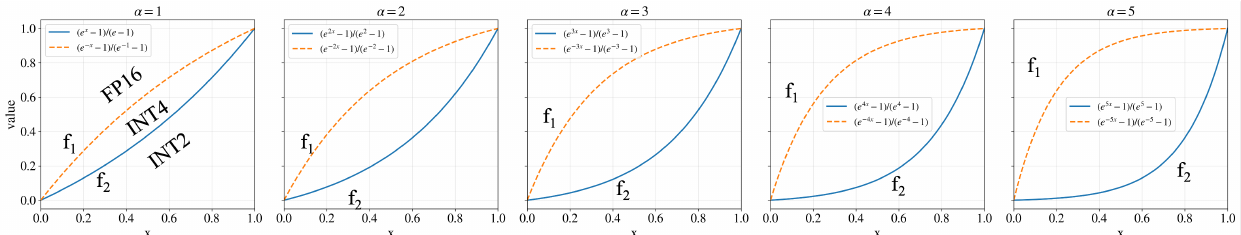}
    \caption{The function plot of $f_1(x)$ and $f_2(x)$ with different $\alpha$.}
    \vspace{-7mm}
    \label{fig: function plot}
    
\end{figure}

We further evaluate the model’s accuracy and token generation latency under different threshold parameters. As shown in Figure \ref{fig: thre_para}, as $\alpha$ increases, both the latency and the model’s accuracy first decrease and then increase. To achieve a good balance between latency and accuracy, we choose 2 as the threshold parameter value used in our main results.
We also analyze the distribution of tokens with different bit-widths under various threshold parameters, as shown in Figure \ref{fig: thre_para_proportion}. As $\alpha$ increases, the numbers of INT2 and FP16 tokens gradually decrease, while the number of INT4 tokens gradually increases. This is consistent with expectations, because the plots of the two threshold functions $f_1(x)$ and $f_2(x)$ (see Equation \ref{eq: f1} and Equation \ref{eq: f2})
show that as $\alpha$ grows, the middle region between the two curves becomes wider (as illustrated in Figure \ref{fig: function plot}), and the tokens falling into this middle region are exactly those that are quantized to INT4.

\begin{table}[t]
\centering
\begin{minipage}[t]{0.45\textwidth}  
\centering
\caption{The configuration of each method in the ablation study. ``$\times$'' means the method doesn't have that module, while ``$\checkmark$'' means the method has that module.}
\resizebox{1\textwidth}{!}{
\begin{tabular}{c|ccc}
\toprule
Method & Module I & Module II & Module III \\
\midrule
A & $\times$ & $\checkmark$ & $\checkmark$ \\
B & $\checkmark$ & $\times$ & $\checkmark$ \\
C & $\checkmark$ & $\checkmark$ & $\times$ \\
\midrule
WindowQuant & $\checkmark$ & $\checkmark$ & $\checkmark$ \\
\bottomrule
\end{tabular}}
\label{tab: ablation method}

\end{minipage}
\hfill                   
\begin{minipage}[t]{0.5\textwidth}
\centering
\caption{The accuracy, memory usage, and decoding throughput of different methods in the ablation study.}
\resizebox{1\textwidth}{!}{
\begin{tabular}{c|ccc}
\toprule
Method & \makecell{Accuracy \\ (\%)} & \makecell{Memory Usage \\ (GB)} & \makecell{Decoding Throughput \\ (tokens/s)} \\
\midrule
Full Precision & 58.1 & 47.7 & 444 \\
\midrule
A & 57.2 & 43.6 & 1276 \\
B & 55.2 & 43.7 & 1250\\
C & 58.4 & 44.7 & 500 \\
\midrule
WindowQuant & 58.4 & 43.7 & 1250  \\
\bottomrule
\end{tabular}}
\label{tab: ablation study}

\end{minipage}
\end{table}

\begin{table}[t]
\centering
\begin{minipage}[t]{0.48\textwidth}  
\centering
\caption{The impact of operator fusion on latency.}
\resizebox{1\textwidth}{!}{
    \begin{tabular}{c|cc}
    \toprule
        Method & Attention Latency & Decoding Latency \\
        \midrule
        Operator Fusion & 0.82 ms & 1640 ms \\
        w/o Operator Fusion & 1.25 ms & 2198 ms  \\
        \bottomrule
    \end{tabular}}
    \label{tab:operator fusion}

\end{minipage}
\hfill                   
\begin{minipage}[t]{0.48\textwidth}
\centering
\caption{Comparison of two different quantization methods.}
\resizebox{1\textwidth}{!}{
\begin{tabular}{c|cc}
\toprule
    Quantization Method & Accuracy & Quantization Metadata \\
    \midrule
    Per-tensor & 64.1 \% & 61.4 MB \\
    Group-wise & 58.6 \% & 0.2 MB \\
     \bottomrule
\end{tabular}}
\label{tab:ablation quantization}

\end{minipage}
\vspace{-3mm}
\end{table}

\begin{table}[]
\caption{Comparison of different similarity functions.}
    \centering
    \begin{tabular}{c|cc}
    \toprule
       Similarity Function & Latency & Accuracy  \\
    \midrule
        Pearson Correlation & 55 ms & 63.2 \%  \\
        Euclidean Distance & 49 ms & 61.4 \% \\
        Cosine Similarity & 48 ms & 64.1 \% \\
    \bottomrule
    \end{tabular}
    \vspace{-3mm}
    \label{tab:sim function}
\end{table}

\subsection{Ablation Study}
Finally, we conduct an ablation study to evaluate the importance of the three modules in WindowQuant. Table \ref{tab: ablation method} shows the configuration of the three methods we test. ``Module I'' means the first window with fixed precision. ``Module II'' means the quantization configuration determination. ``Module III'' means the KV cache window reordering.  Here, ``w/o Module II'' refers to randomly assigning the quantization configuration for each window. We set the batch size to 60, extract 5 frames per video, and test those methods on the EgoSchema datasets with LLaVA-OneVision-Qwen2-7B model. Table \ref{tab: ablation study} shows the ablation study results. From the table, we can see that removing the first-window fixed precision leads to a slight drop in model accuracy, while latency and memory usage are slightly improved. This is because, in this case, the first window may also be quantized, increasing the number of quantized tokens. However, since the first window is crucial, quantizing it reduces the model’s accuracy.
Removing the quantization configuration determination results in a more noticeable drop in accuracy, but latency and memory usage remain unchanged. This indicates that our quantization configuration determination indeed plays a key role in identifying important tokens.
When KV cache window reordering is removed, the model’s accuracy remains unchanged, which is consistent with our conclusion that KV cache window reordering does not affect accuracy. However, memory usage increases a little and decoding throughput drops significantly because having KV cache segments with mixed precisions in physically contiguous memory reduces hardware efficiency.

We also evaluate the impact of operator fusion on latency, as summarized in Table \ref{tab:operator fusion} (batch size = 20, number of frames = 5). With operator fusion enabled, the per-layer attention computation latency for decoding a single token is reduced by 0.43 ms, leading to an overall reduction of 558 ms in total decoding latency.

We further set the batch size to 1 and the number of frames to 100 to evaluate the effects of per-tensor versus group-wise quantization, as well as different similarity computation functions. As shown in Table \ref{tab:ablation quantization}, compared with per-tensor quantization, group-wise quantization slightly increases the memory footprint of quantization metadata (by approximately 61.2 MB, which is negligible compared to the overall memory savings brought by quantization), while achieving an accuracy improvement of about 5.5\%. Therefore, we consider group-wise quantization to be highly worthwhile. Table \ref{tab:sim function} presents a comparison of different similarity computation functions. We evaluate three functions—pearson correlation, euclidean distance, and cosine similarity—for similarity calculation, and test their computation latency and task accuracy (batch size = 1, number of frames = 100). The results show that cosine similarity achieve the lowest latency and the highest accuracy. Therefore, we choose cosine similarity as our similarity function.

\section{Related Works}

\subsection{KV cache Compression}
Past researchers have proposed various methods to compress the KV cache. Some have suggested pruning the attention module within the KV cache \cite{shazeer2019fast,ainslie2023gqa,cai2024medusa}. For example,  Shazeer \cite{shazeer2019fast} proposed MQA, where all attention heads share a single set of KV parameters. Some have proposed evicting unimportant tokens from the KV cache \cite{xiao2023efficient,zhang2024h2o,han2023lm}. For instance, Xiao et al. \cite{xiao2023efficient} introduced streamingLLM, which retains only the most recent sliding window of KV cache along with the initial tokens. Zhang et al. \cite{zhang2024h2o} proposed Q-Hitter, which evaluates the importance of attention based on the sum of attention scores across each token's column in the attention matrix. Others have approached the problem from a system perspective \cite{dao2022flashattention, kwon2023efficient,dao2023flashattention}. For example, Kwon et al. \cite{kwon2023efficient} introduced PageAttention, which applies the memory paging concept from traditional operating systems by mapping logically contiguous KV cache pages to physically non-contiguous pages through a page table. Dao et al. \cite{dao2022flashattention} proposed FlashAttention, which rewrites certain complex operators to be processed as much as possible within the GPU's high-speed SRAM. However, these methods are not as easy to implement as quantitative methods, and their optimization results are also not as effective as those achieved by quantitative methods.

\subsection{Quantization}
Quantization is also an effective method to optimize VLMs' memory footprint and inference latency. The quantization work initially focused on quantizing weights and activations \cite{xiao2023smoothquant, frantar2022gptq,dettmers2022gpt3,ge2023model,dettmers2023spqr,lin2024awq}. For example, Xiao et al. \cite{xiao2023smoothquant} proposed SmoothQuant, which scales weights that are easy to quantize and activations that are difficult to quantize separately.
However, during inference with a long input sequence, the KV cache often becomes larger than the weights and activations. Therefore, the quantization work has been extended to the KV cache as well \cite{sheng2023flexgen,zhao2024atom,liu2024kivi,lin2024qserve}. For instance, Zhao et al. \cite{zhao2024atom} introduced Atom, which performs group quantization of the KV cache to low bits, where each group has independent quantization parameters. Liu et al. \cite{liu2024kivi} proposed KIVI, which applies per-channel quantization to the K cache while keeping the original per-token quantization for the V cache. However, these quantization methods that uniformly quantize tokens to the same bit-width cannot maintain model accuracy effectively. Other researchers have proposed mixed-precision quantization methods to process important tokens \cite{kim2023squeezellm,yang2024no,hooper2024kvquant,dong2024qaq}. For example, Kim et al. \cite{kim2023squeezellm} proposed SqueezeLLM, which divides the weights into dense matrices without outliers and sparse matrices containing outliers, then applies low-bit quantization to the dense matrices while keeping the outliers in FP16 precision. Yang et al. \cite{yang2024no} proposed MiKV, which identifies unimportant tokens based on attention scores but quantizes them instead of evicting them. Hooper et al. 
 \cite{hooper2024kvquant} introduced KVQuant, which uses a custom data type called nuqX to represent the mixed-precision quantized KV cache. These methods use token-level quantization search, which is very time-consuming, and they have not adequately addressed the hardware inefficiency issues brought by mixed-precision quantization.

\section{Conclusion}
This paper introduces WindowQuant, a window-adaptive mixed-precision KV cache quantization method for VLM inference. WindowQuant uses a window-level quantization search module to determine the bit-width configuration of visual token KV cache windows. It also contains a window-level KV cache computation module, reordering these KV cache windows to avoid hardware inefficiency. Extensive experiments on multiple datasets and models demonstrate that WindowQuant outperforms SOTA VLM models and KV cache quantization methods.

\bibliographystyle{ACM-Reference-Format}
\bibliography{sample-base}

\end{document}